\pdfoutput=1

\documentclass[11pt]{article}

\usepackage[final]{acl}

\usepackage{times}
\usepackage{latexsym}
\usepackage{amsmath}
\usepackage{comment}
\usepackage{subcaption}
\usepackage{multirow}
\usepackage[T1]{fontenc}

\usepackage[utf8]{inputenc}

\usepackage{microtype}

\usepackage{inconsolata}

\usepackage{graphicx}
\usepackage{booktabs}
\usepackage{tabularx}
\usepackage[flushleft]{threeparttable}  

\title{Formalizing Style in Personal Narratives}

\author{
  \textbf{Gustave Cortal\textsuperscript{1,2}},
  \textbf{Alain Finkel\textsuperscript{1}},
\\
  \textsuperscript{1}Université Paris-Saclay, CNRS, ENS Paris-Saclay, LMF, 91190, Gif-sur-Yvette, France\\
  \textsuperscript{2}Université Paris-Saclay, CNRS, LISN, 91400, Orsay, France
\\
  \small{
    \{gustave.cortal, alain.finkel\}@ens-paris-saclay.fr
  }
}

\begin{document}
\maketitle
\begin{abstract}
Personal narratives are stories authors construct to make meaning of their experiences. Style, the distinctive way authors use language to express themselves, is fundamental to how these narratives convey subjective experiences. Yet there is a lack of a formal framework for systematically analyzing these stylistic choices. We present a novel approach that formalizes style in personal narratives as patterns in the linguistic choices authors make when communicating subjective experiences. Our framework integrates three domains: functional linguistics establishes language as a system of meaningful choices, computer science provides methods for automatically extracting and analyzing sequential patterns, and these patterns are linked to psychological observations. Using language models, we automatically extract linguistic features such as processes, participants, and circumstances. We apply our framework to hundreds of dream narratives, including a case study on a war veteran with post-traumatic stress disorder. Analysis of his narratives uncovers distinctive patterns, particularly how verbal processes dominate over mental ones, illustrating the relationship between linguistic choices and psychological states.
\end{abstract}

\section{Introduction}

\begin{table}[!ht]
  \centering
  \small
  \renewcommand{\arraystretch}{1.1}
  \begin{threeparttable}
    \caption{Illustrative pipeline for our sequence-based framework. We first segment “I wake in a dark room. I feel a cold wind. I tell myself to move.” into clauses, then identify features such as processes and participants for each clause. Each narrative is mapped to a symbolic sequence using an alphabet based on extracted features.}
    \label{tab:example}
    \begin{tabular}{lll}
      \toprule
      \textbf{Clause} & \textbf{Process} & \textbf{Participants} \\
      \midrule
      I wake in a dark room         & Action (\textbf{a})  & Actor \\
      I feel a cold wind            & Mental (\textbf{m})  & Senser,\\
                                            &             & Phenomenon \\
      I tell myself to move         & Verbal (\textbf{v})  & Sayer,\\
                                            &             & Recipient \\
      \bottomrule
    \end{tabular}

    \begin{tablenotes}[flushleft]
      \footnotesize
      \item \textbf{Sequence:} $amv$\quad|\quad
            \textbf{Substrings:} \{am, mv\}
    \end{tablenotes}
  \end{threeparttable}
\end{table}

As humans, we use narratives to express our representations of reality and make sense of the world \cite{brunerActsMeaning1990}. Through written narratives, authors communicate events and reveal their unique ways of perceiving and interpreting reality. Personal narratives, the self-crafted stories authors create to make sense of their experiences, are rich sources for understanding how authors shape their identity and meaning-making \cite{labovNarrativeAnalysisOral1997}. 

Style has long been recognized as a key element in how authors express themselves through language. In everyday usage, style refers to a distinctive manner of expression\footnote{\url{https://merriam-webster.com/dictionary/style}}. While this general definition encompasses all aspects of personal expression, our work narrows the focus to specifically examine how individuals linguistically encode their subjective experiences in personal narratives. Our approach complements existing works in stylometry \cite{nealSurveyingStylometryTechniques2017} and stylistics \cite{walesDictionaryStylistics2014a} by providing a formal framework grounded in systemic functional linguistics \cite{hallidayIntroductionFunctionalGrammar2014}, which views language as a system of options for achieving communicative goals. This perspective aligns with how authors construct personal narratives, as they select from available linguistic resources to convey their subjective experience.

Scholarly investigations have long explored personal modes of reasoning and expression. For example, Husserl’s phenomenology examined how intentional structures of consciousness shape lived experience \cite{husserl2012ideas}, Hadamard analyzed the cognitive processes underlying mathematical creativity \cite{hadamard1945essay}, and Dilts modelled the problem-solving strategies of figures such as Leonardo da Vinci and Sherlock Holmes \cite{dilts1994strategies}. In his philosophy of style, Gilles-Gaston Granger defined style as the individual response to the difficulties faced by any structuring endeavor \cite{granger1968essai}. While these qualitative frameworks richly describe personal modes of thought, they do not provide operational tools for capturing how such styles manifest in linguistic form.

A formalization of the intuitive notion of style in personal narratives is lacking. In therapeutic settings, where narrative reconstruction can facilitate healing \cite{whiteNarrativeMeansTherapeutic1990}, formalization could enable more precise identification of linguistic patterns associated with psychological states and support targeted interventions. 

\citet{TellierFinkel95} propose a formal framework to characterize individual patterns between texts written by the same author. They propose a definition of linguistic style as lexical and syntactical patterns in the expressions of an intention. Drawing on these ideas, we develop a sequence-based framework to analyze how personal narratives convey subjective experience. Table \ref{tab:example} shows an illustrative pipeline for our framework. We represent narratives as sequences of linguistic choices, identifying recurring patterns in how authors encode their experiences. We do not claim to capture the full complexity of subjective experience in text. We aim to create a simple, accessible framework that researchers can use across disciplines and build upon in future studies. Our key contributions are:

\begin{enumerate}
    \item A sequence-based framework defining style as patterns in sequences of linguistic choices grounded in systemic functional linguistics;
    \item A methodology for automatically identifying patterns using sequence analysis;
    \item A case study on dream narratives, showing how the analysis of patterns can reveal psychological insights and support therapeutic applications.
\end{enumerate}

Our research task is to formalise style in personal narratives as patterns of linguistic choices that encode subjective experience, and to demonstrate that this formalisation yields interpretable empirical findings. Section \ref{categorization} shows our categorization of linguistic features grounded in functional linguistics and how these features can be organized as sequences of symbols. Section \ref{redundancies} presents methods for analyzing patterns in these sequences. Section \ref{case_study} demonstrates our framework through a case study on dream narratives. Finally, we conclude and propose many applications (\textit{e.g.}, authorship profiling, style-conditioned narrative generation) and theoretical extensions (applying methods from complexity science and formal language theory).

\begin{table*}[!htb]
    \centering
    \begin{tabular}{p{4cm}|p{11cm}}
        \hline
        \textbf{Processes} & \textbf{Examples} \\ \hline
        \textit{Action}: actions and events in the physical world. &
        [He]$_{\text{Actor}}$ [\textbf{takes}]$_{\text{Action}}$ [the valuable]$_{\text{Affected}}$ \newline
        
        [Members of my cult]$_{\text{Actor}}$ [\textbf{have made]}$_{\text{Action}}$ [1500 euros]$_{\text{Result}}$ \newline
        
        [I]$_{\text{Actor}}$ [\textbf{give}]$_{\text{Action}}$ [her]$_{\text{Recipient}}$ [a chance]$_{\text{Range}}$ \\ \hline
        
        \textit{Mental}: internal experiences such as thoughts, perceptions, and feelings. &
        [We]$_{\text{Senser}}$ [\textbf{believe}]$_{\text{Mental}}$ [women are the leaders of change]$_{\text{Phenomenon}}$ \newline
        
        [The moon]$_{\text{Senser}}$ [\textbf{sees}]$_{\text{Mental}}$ [the earth]$_{\text{Phenomenon}}$ \newline
        
        [He]$_{\text{Senser}}$ [\textbf{disliked}]$_{\text{Mental}}$ [Gilbert's writing]$_{\text{Phenomenon}}$ \\ \hline
        
        \textit{Verbal}: acts of communication. &
        [David]$_{\text{Sayer}}$ [\textbf{said}]$_{\text{Verbal}}$ [``the corrupt, criminals and money launderers'']$_{\text{Verbiage}}$ \\ \hline
        
        \textit{State}: states of being, having, or existence. &

         There [\textbf{was}]$_{\text{Existential}}$ [a swimming pool]$_{\text{Existent}}$ \newline
        
        [John]$_{\text{Carrier}}$ [\textbf{is}]$_{\text{State}}$ [an interesting teacher]$_{\text{Attribute}}$ \newline
        
        [Hadrian's Wall]$_{\text{Possessor}}$ [\textbf{has}]$_{\text{State}}$ [something for everyone]$_{\text{Possessed}}$ \\ 
    \end{tabular}
    \caption{Processes with their participants.}
    \label{tab:process_participants}
\end{table*}

\section{Categorizing Linguistic Features\label{categorization}}

What linguistic features are relevant for analyzing the communication of a subjective experience? This section presents our categorization of linguistic features based on functional linguistic theories. In these theories, language is seen primarily as a means to achieve communicative goals. Functional linguistics often draws on various disciplines, such as sociology, psychology, and cognitive science, to explore how language functions. We rely on a fundamental function of language: the communication of a subjective experience. It is through this notion that we define our concept of style. We introduce the various functional linguistic theories related to this function, then we present our categorization of features based on systemic functional linguistics. 

\paragraph{Functions of language} Karl Bühler's organon model represents one of the earliest attempts to categorize language functions \cite{buhlerTheoryLanguageRepresentational1990}. He identified three fundamental functions: \textit{representation} (encoding external reality), \textit{expression} (revealing internal states), and \textit{appeal} (influencing receivers of messages). The expressive function is particularly relevant to our analysis, as it specifically addresses how language manifests the internal state of the writer through linguistic choices, independent of both representational content and intended receiver effects. Roman Jakobson later expanded this model into a six-function framework \cite{jakobsonLinguisticsPoetics2010}.

\paragraph{Systemic functional linguistics} Sharing conceptual roots with Bühler's and Jakobson's models of language, systemic functional linguistics, developed by Michael Halliday, offers a framework for analyzing how language functions as a meaning-making resource \cite{hallidayIntroductionFunctionalGrammar2014}. Rather than viewing language as a set of abstract rules, systemic functional linguistics examines it as a social semiotic system where meaning emerges through contextual linguistic choices. This approach is particularly valuable for analyzing personal narratives, as it provides tools for understanding how authors select and combine linguistic resources to construct meaning from their experiences. Systemic functional linguistics identifies three metafunctions: interpersonal (how language is used to build and maintain social relationships), textual (how information is organized to create coherent messages), and ideational (how language represents experience through the transitivity system described below). Our framework particularly leverages the transitivity system, analyzing how authors encode their experiences through processes, participants, and circumstances.

\paragraph{Processes, participants, and circumstances} The transitivity system, which shares similarities with Lucien Tesnière's concept of valency \cite{tesniereElementsStructuralSyntax2015}, provides a framework for analyzing how clauses represent experiences through the interaction of processes, participants, and circumstances. Processes (realized through verbal phrases), participants (realized through noun phrases), and circumstances (realized through adverbial groups or prepositional phrases) form the core of experiential representation. Inspired by \citet{banksSystemicFunctionalGrammar2019b}, Table \ref{tab:process_participants} presents our categorization of process types. Although there exists a finer categorization of processes (for example, the division of mental processes into cognitive, affective, and perceptive processes), we limit ourselves to a general categorization to save space. The definitions of participants and examples of circumstances can be found in Tables \ref{tab:process_types_no_examples} and \ref{tab:circumstances} (Appendix \ref{sec:appendix}).

We define \textit{style} as patterns in choices within a system of linguistic features inspired by the transitivity system in systemic functional linguistics. We now need to explain what is a choice. To account for a subjective experience, an author makes a series of choices within a system of features. 


\section{Capturing Sequential Patterns\label{redundancies}}

\subsection{Formalizing Choice\label{formalization_choice}}

\paragraph{Grammar as a system of meaningful choices} Systemic functional linguistics views grammar as a network of meaningful choices \cite{hallidayIntroductionFunctionalGrammar2014}. This paradigmatic dimension emphasizes how linguistic units are selected from available options \cite{Saussure1916}. When constructing clauses, authors select features from the transitivity system to encode their experience, making choices that fulfill specific communicative functions.

\paragraph{Formal representation of feature systems} We represent each system of linguistic features as a finite set $\Sigma$ containing available choices, also called an alphabet in formal language theory \cite{pinHandbookAutomataTheory2021}. For example, the alphabet of process types is $\Sigma_\text{process} = \{action, mental, verbal, state\}$. Using this alphabet, we can construct a word $w$ with five letters, such as $w = action.mental.action.mental.state$ where $.$ is the concatenation operator. Multiple alphabets can be combined through cartesian products, allowing the representation of simultaneous choices across different linguistic dimensions. For example, we can have a finer-grained analysis with processes along with their tense and aspect: $\Sigma = \Sigma_\text{process} \times \Sigma_\text{tense} \times \Sigma_\text{aspect}$. Our case study on dream narratives is a simple proof of concept where we only focus on process types. Future works could incorporate more features, hence a larger alphabet, for a finer-grained analysis. 

\subsection{Sequence Analysis}

\paragraph{Substrings and subsequences} We analyze sequences to identify two types of recurring patterns: substrings and subsequences. A \emph{substring} is any contiguous block of symbols occurring within a longer string. A \emph{subsequence} is obtained by deleting zero or more symbols while preserving their relative order, so contiguity is not required. For example, $action.action$ is a subsequence of $action.mental.state.action$ but not a substring, whereas $action.mental$ is both a substring and a subsequence. Both notions come from the computer science terminology and underpin a large family of string-processing algorithms \cite{knuthFastPatternMatching1977}. In formal language theory, substrings and subsequences are called \emph{factors} and \emph{subwords} \cite{berstelOriginsCombinatoricsWords2007}. 

Our study focuses on substrings, but our framework can be easily extended to consider subsequences. Identifying recurring substrings such as $action.mental$ within words can help detect patterns relevant to an author's style. For example, an author may tend to follow an action process with a mental process. By applying this framework to larger corpora, we can explore how authors structure their narratives, revealing the patterns that shape how subjective experience is communicated.

\paragraph{Analogy with computational biology} As we aim to identify patterns in sequences of symbols, we can draw inspiration from concepts and methods developed in various fields that perform sequence analysis, such as computational biology \cite{watermanIntroductionComputationalBiology1995}. For example, studying DNA or protein sequences in biology has led to the development of algorithms and similarity measures, notably through sequence alignment and complexity measures. These tools allow for identifying recurring patterns within sequences, quantifying the proximity between biological entities, and the inference of functional or evolutionary relationships. By analogy, if we replace nucleotides (\textit{A}, \textit{C}, \textit{G}, \textit{T}) with our linguistic features, we can employ these techniques to detect recurring patterns, measure the stylistic similarity between different narratives, and highlight patterns specific to certain authors.

\paragraph{Similarity measures} When comparing narratives, we focus on their stylistic patterns rather than their specific content by analyzing their associated sequences. While we could consider two narratives equal only if they have exactly the same sequence, this criterion is too strict for practical analysis. Instead, we need a more flexible way to measure how similar two sequences are.
We propose using cosine similarity, which treats sequences as vectors in a high-dimensional space and measures the angle between them. For any two sequences $s_1$ and $s_2$, we first identify all possible substrings (contiguous subsequences) up to a certain length. We then count how frequently each substring appears in each sequence to create frequency vectors. The similarity is calculated as:
\[
cos(s_1, s_2) = \frac{\sum_i x_i y_i}{\sqrt{\sum_i x_i^2} \sqrt{\sum_i y_i^2}}
\]
where $x_i$ and $y_i$ represent the number of times a particular substring appears in $s_1$ and $s_2$ respectively. This measure ranges from 0 (completely different) to 1 (identical), with higher values indicating greater similarity.
For example, consider two sequences of process types: $s_1 = amvma$ and $s_2 = amma$ (where $a=action$, $m=mental$, $v=verbal$). Their substrings of length one and two are: $s_1: \{a:2, m:2, v:1, am:1, mv:1, vm:1, ma:1\}$ and $s_2: \{a:2, m:2, am:1, mm:1, ma:1\}$. The cosine similarity between these sequences is 0.836. This high similarity reflects that they share many common substrings and have similar distributions of process types, even though $s_1$ contains a verbal process not present in $s_2$. Other similarity measures could be used, such as the Jaccard similarity or Euclidean distance.

\paragraph{Clustering} We aim to find groups of narratives that reflect similar subjective experiences encoded in their patterns. We apply unsupervised clustering methods, particularly hierarchical agglomerative clustering \cite{wardHierarchicalGroupingOptimize1963}, to automatically group similar sequences. This method constructs a hierarchy (or dendrogram) of clusters in a bottom-up manner. There is no need to specify the number of clusters as the methods produce a full hierarchy, allowing the analyst to choose a clustering level that best suits the needs. By examining the dendrogram, one can identify the number of clusters that reflect meaningful stylistic distinctions. For instance, one might observe that certain narratives cluster together at a coarse level while finer-grained distinctions emerge deeper in the dendrogram. 

\paragraph{Representative sequences} One challenge in interpreting clustering results is to summarize each cluster in a way that highlights its most characteristic features. Representative sequences can summarize what stylistically unites the narratives in a cluster and offer a starting point for qualitative interpretation. To select representative sequences, we compute the sequence with the smallest average distance to all other sequences in the cluster. 

In the next section, we apply sequence analysis methods to identify recurring patterns in dreams. 

\section{Case Study on Dream Narratives\label{case_study}}

To demonstrate the applicability of our framework, we perform a case study on dream narratives. By leveraging language models to automatically extract linguistic features at scale, we map each narrative into a sequence and uncover recurring patterns through sequence analysis. The patterns we identify offer insights into the psychological dimensions of how people articulate their experiences.

\subsection{Data and Methodology}

\paragraph{DreamBank corpus} We apply our framework to dream narratives as they possess a narrative structure and represent intentional attempts to communicate subjective experience. Our analysis uses DreamBank, a corpus of thousands of dream narratives collected in the United States for scientific research. This English-language corpus has been employed in quantitative dream analysis \cite{domhoffStudyingDreamContent2008a}. The corpus is provided for academic research in dream and narrative analysis, and our usage strictly adheres to this intended purpose. From this corpus, we analyze five series of dreamers: \textit{blind} (long-term blind dreamers, n=361), \textit{ed} (a widower, n=139), \textit{izzy} (a teenager, n=1091), \textit{merri} (an artist, n=202), and \textit{viet} (a Vietnam War veteran with post-traumatic stress disorder, n=566). A description of each series is accessible on the DreamBank website\footnote{\url{https://www.dreambank.net/grid.cgi}}. 

While this paper focuses on analyzing the \textit{viet} series, we make the sequences of other series available. We construct a \textit{norm} to establish a comparative baseline to compare how each series deviates from a hypothetical average dreamer. This norm comprises a random sample of ten narratives from each series of DreamBank, resulting in a collection of 720 dream narratives. 

\paragraph{Methodology} Prior research has demonstrated the effectiveness of language models in analyzing dream narratives for character and emotion prediction \cite{bertolini2023automatic,cortal-2024-sequence}. Previous studies have also shown that systemic functional linguistics can be valuable for analyzing text corpora \cite{banksSystemicFunctionalLinguistics2002a}. Our work represents the first attempt to automate analysis from systemic functional linguistics using language models, focusing on the ideational function, particularly the transitivity system. We automatically extract processes, participants, and circumstances from dream narratives. This automation eliminates the costly and time-consuming manual extraction process that requires trained annotators and annotation campaigns.

 Our methodology begins with dividing dream narratives into sentences using SpaCy segmentation model\footnote{\textit{en\_core\_web\_trf-3.8.0} described at \url{https://spacy.io}}. We then separate sentences into clauses using a language model with in-context examples \cite{dongSurveyIncontextLearning2024}. We do not differentiate between dependent and independent clauses. The next step involves extracting linguistic features using a language model with in-context examples from systemic functional linguistics books \cite{banksSystemicFunctionalGrammar2019b, hallidayIntroductionFunctionalGrammar2014}. We use Llama 3.1 8B Instruct, a instruction tuned auto-regressive language model with 8-billion parameters \cite{grattafioriLlama3Herd2024}. This model was the best-performing 8B open‑weights model at the time of our experiments\footnote{\url{https://huggingface.co/spaces/HuggingFaceH4/open\_llm\_leaderboard}}. We run our experiments using a Tesla V100 32GB for 80 hours. 
 
 As our sequence analysis depends on reliable extraction of linguistic features, we conducted a quantitative validation of the language‑model annotator. We randomly sampled 50 clauses from the examples in \citet{banksSystemicFunctionalGrammar2019b} and \citet{hallidayIntroductionFunctionalGrammar2014}. These clauses are considered gold‑standard because they are fully analysed in the textbooks. Predicted processes, participants and circumstances were all correctly matched to the references.
 
 We make our prompts and all extracted sequences available online for research use\footnote{\url{https://github.com/gustavecortal/formalizing-style-in-personal-narratives}}. We acknowledge that our automatic extraction may contain errors, such as misidentifying a mental process as a verbal process. Future work could evaluate our models against more expert annotations. Our framework benefits from the continuous improvement of language models. Llama 3.1 8B can be replaced with better language models in the future without altering the essence of our approach.

\subsection{Analysis\label{sec:analysis}}

We apply sequence analysis methods to identify recurring patterns in sequences of choices extracted from dream narratives. We aim to identify patterns characterizing how subjective experience is communicated in dream narratives. 

To demonstrate the effectiveness of our framework and provide a simple case, we focus solely on the choice of processes. However, our approach can consider more complex configurations of linguistic features to identify finer-grained patterns. Our analysis focuses on the distribution of substrings and representative sequences of the war veteran with post-traumatic stress disorder. We discuss the psychological implications of our findings.

\begin{figure}[!htb]
     \centering
     \begin{subfigure}[t]{\linewidth}
         \centering
         \includegraphics[scale=0.28]{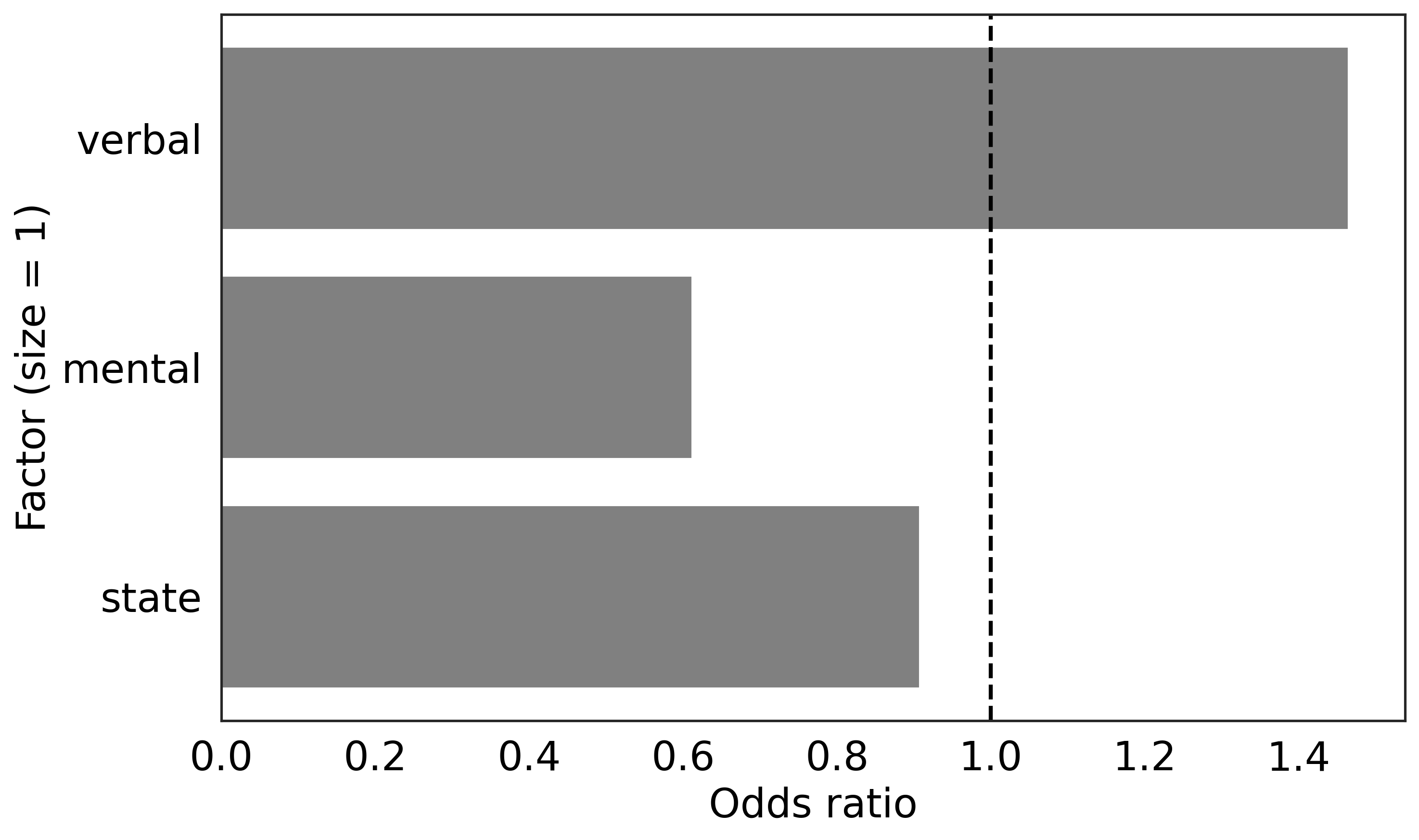}
         \caption{Size 1.}
         \label{fig:viet_odds1}
     \end{subfigure}
     \medskip
     \begin{subfigure}[t]{\linewidth}
         \centering
         \includegraphics[scale=0.28]{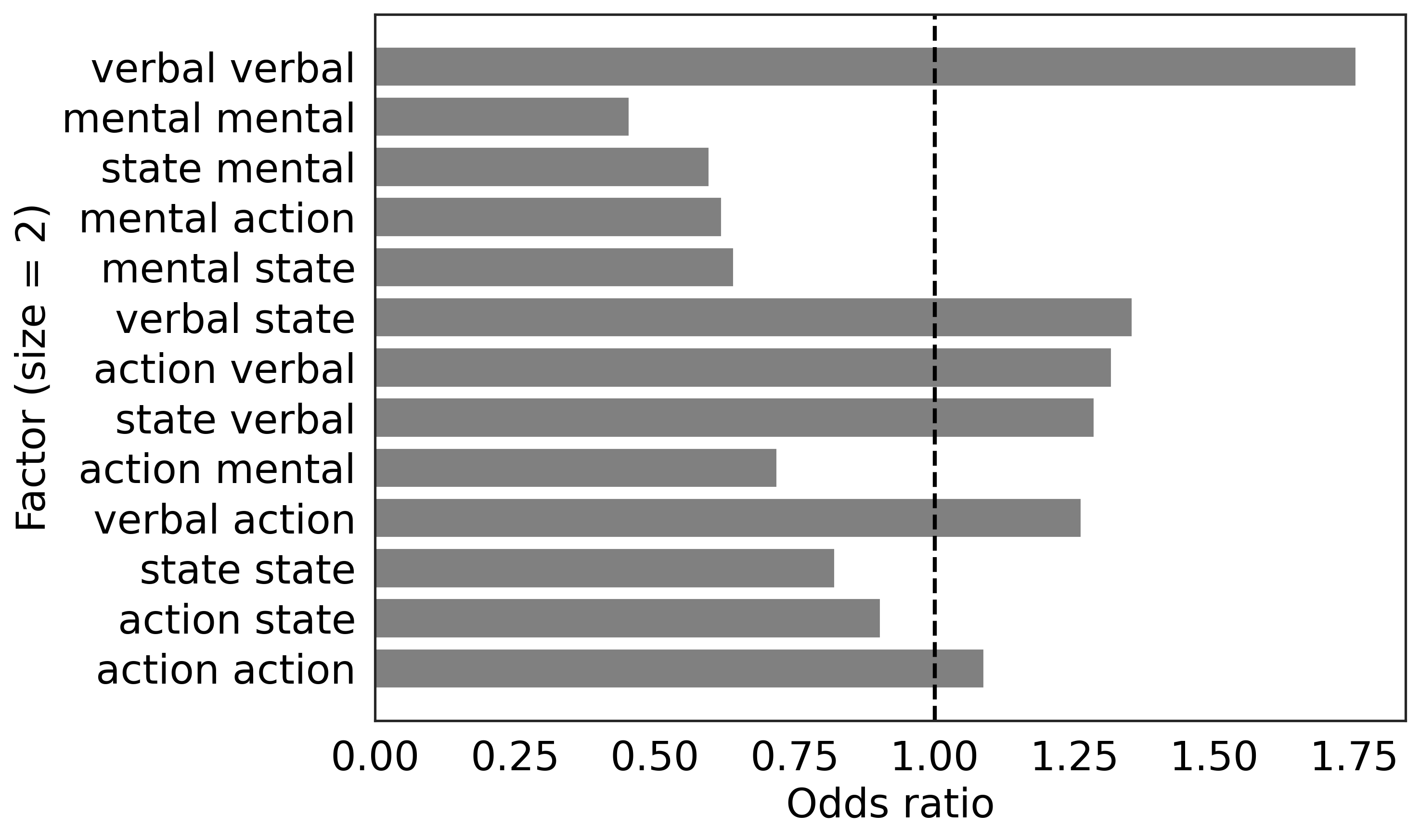}
         \caption{Size 2.}
         \label{fig:viet_odds2}
     \end{subfigure}
         \begin{subfigure}[t]{\linewidth}
         \centering
         \includegraphics[scale=0.28]{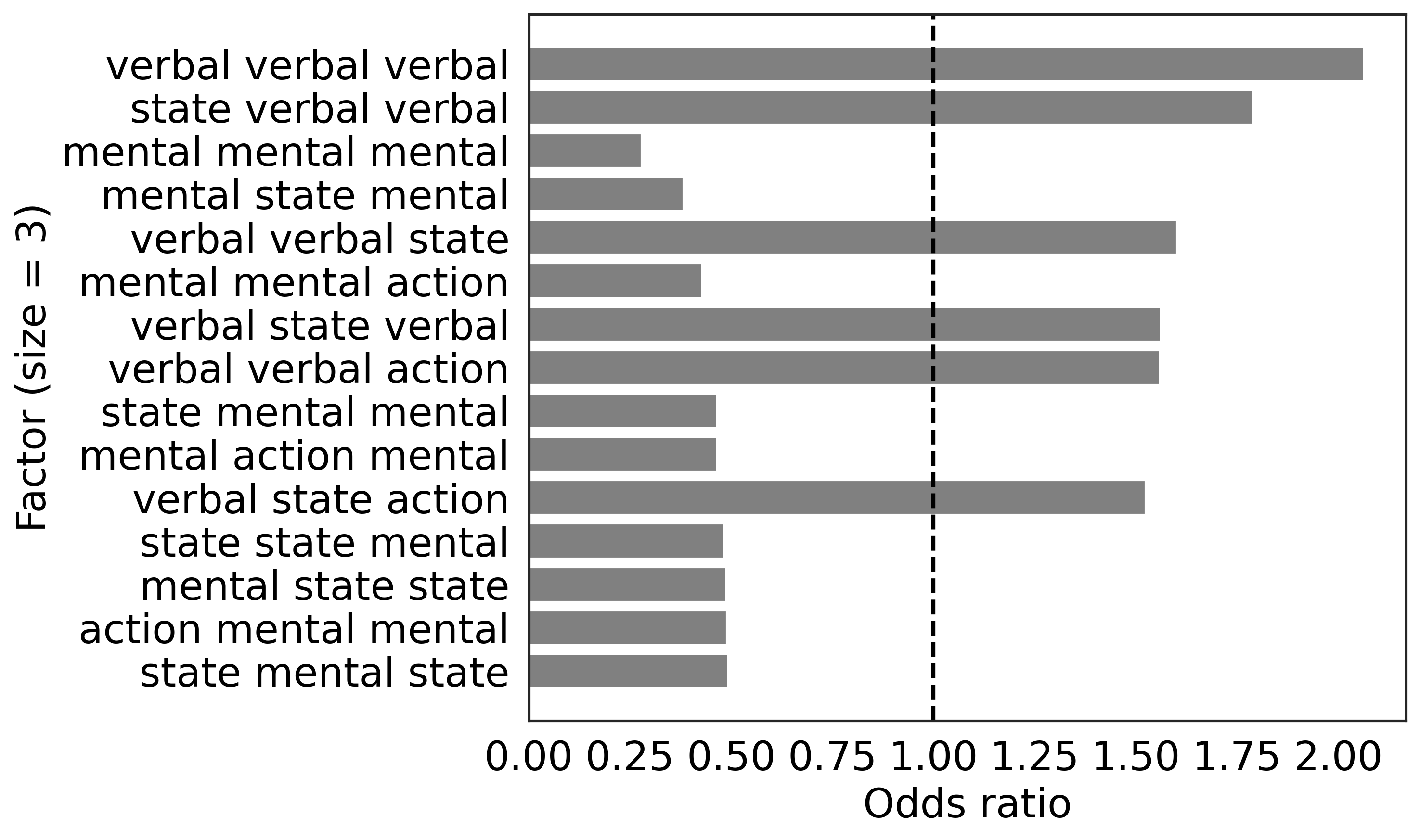}
         \caption{Size 3.}
         \label{fig:viet_odds3}
     \end{subfigure}
        \caption{Top substring between \textit{viet} and \textit{norm}.}
        \label{fig:viet_odds}
\end{figure}

\paragraph{Distribution of substrings} We compare the distribution of substrings between a series of dreamers and an average dreamer (\textit{norm}). To save space, we only compare substrings of sizes one, two, and three. In Figure \ref{fig:viet_odds}, we take \textit{viet} as an example. Comparisons with other series are available in Appendix \ref{sec:appendix}. We also report other statistics such as per-series distribution of sequence lengths (Figure \ref{fig:sequence_length}) and number of distinct substrings in the series (Figure \ref{fig:distinct_substrings}). We compare the proportion of sequences containing a given substring between \textit{viet} and \textit{norm} using a Fisher exact test (with Holm–Bonferroni correction)  \cite{fisherDesignExperiments1935}. We focus on the presence or absence of a substring within a sequence: each substring is treated as a categorical variable indicating whether or not it occurs in a sequence. 

 \paragraph{Odds ratio} Figure \ref{fig:viet_odds} reports odds ratios with statistically significant results ($p < 0.05$). The odds ratio measures how much more (or less) likely it is to observe a given substring in one series compared to another. For example, Figure \ref{fig:viet_odds1} shows that the presence of verbal processes is 40\% more likely in the sequences of \textit{viet} than in \textit{norm}, as we have an odds ratio of 1.4. The presence of mental processes is 40\% less likely in the sequences of \textit{viet} than in \textit{norm}, as we have an odds ratio of 0.6. Figures \ref{fig:viet_odds1}, \ref{fig:viet_odds2}, and \ref{fig:viet_odds3} show a preference for \textit{viet} to remain in a verbal process, as indicated by substrings such as \textit{verbal.verbal} and \textit{verbal.verbal.verbal} with high odds ratios (respectively 2.00 and 1.75). We save space by showing only substring sizes up to three. However, we analyze other sizes that reveal interesting patterns. For example, some substrings do not appear in \textit{viet} but appear in \textit{norm}, such as \textit{mental.state.mental.mental}.
 

\begin{figure}[!htb]
    \centering
    \includegraphics[width=0.95\linewidth]{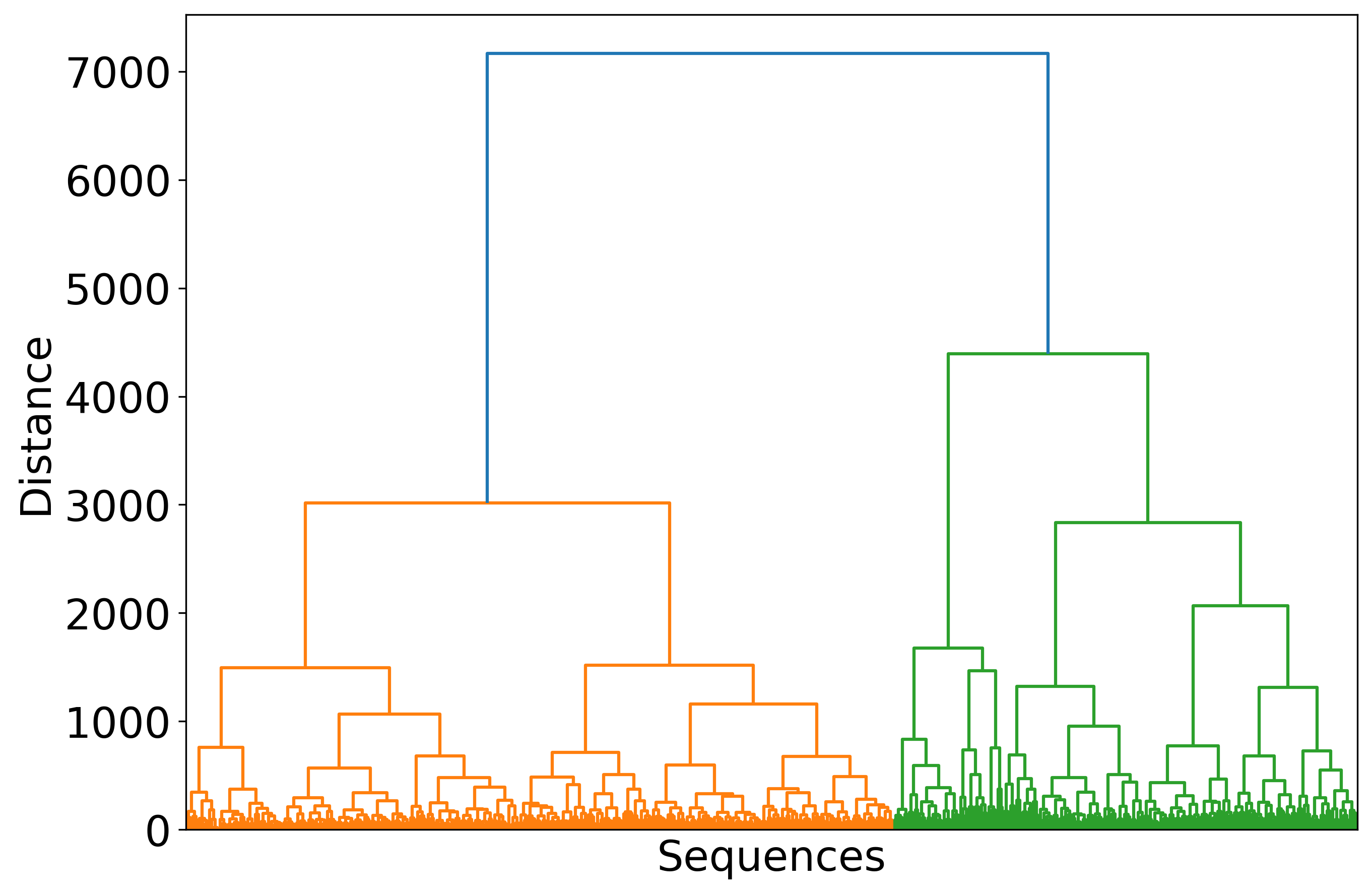}
    \caption{Dendrogram for \textit{viet} based on substrings of size one, two, and three.}
    \label{fig:dendogram}
\end{figure}

\paragraph{Representative sequences\label{representative}} We perform hierarchical agglomerative clustering on the sequences of \textit{viet} to find representative sequences that cover an important portion (80\%) of all sequences while minimizing intra-cluster distances. We use Ward linkage and cosine similarity. The silhouette score, which measures how similar sequences are to their own cluster compared to other clusters, reaches its maximum value with substrings of lengths up to three and when partitioning the dendrogram into two clusters (shown in orange and green in Figure \ref{fig:dendogram}). This suggests two distinct patterns in how the veteran structures his subjective experience in dream narratives. We identified a representative sequence for each cluster by calculating the sequence with the minimal average distance to all other sequences within its cluster. The first representative sequence (\textit{savamasasaaamaaasavvvaaaaaaavssaaaaa}) demonstrates a dominance of action processes (23 occurrences) with minimal mental processes (2) and covers 274 sequences with an average distance of 8 to other sequences in its cluster. The second representative sequence (\textit{sssssavaavssvsavvvvsmasasaasasaamaamvmsss}) reveals a more balanced pattern, featuring both action (13 occurrences) and state processes (16 occurrences) while maintaining relatively few mental processes (4), and covers 179 sequences with an average distance of 21. These patterns suggest that the veteran follows two templates: a highly action-oriented structure or a more varied structure alternating between state and action processes.

\paragraph{Psychological interpretations} We can bridge our analysis with therapeutic goals. Identifying low- or high-odd substrings could serve as diagnostic markers or guide therapy interventions. As therapists and patients collaborate to restructure narratives, certain stylistic elements can help target specific patterns for modification. For instance, mental processes occurring more frequently than average in an author's narratives might indicate an intensified inward focus. Conversely, a relative scarcity of such patterns could suggest a more action-oriented or externally focused stance. In our example, \textit{viet} primarily frames his experiences through action and verbal rather than mental processes. This may align with studies about how combat trauma can affect cognitive and emotional processing \cite{americanpsychiatricassociationDiagnosticStatisticalManual2013}, though further research would be needed to confirm this interpretation. We aim to show that our formalization of style can help authors learn to reframe their experiences, potentially facilitating trauma processing.

Our analysis relies solely on process types to show a simple proof of concept. Future works will include a fine-grained sequence analysis considering processes, participants, and circumstances. These building blocks can be further specified with features with a functional role, such as the duration and completeness of processes (realized by tense and aspect) and the concreteness of participants. 

\section{Related Work}

\paragraph{Systemic Functional Linguistics parsing.}
Early efforts to operationalise Halliday’s transitivity system were almost exclusively rule‐driven \cite{honnibal2010handoff,odonnell2016handbook}. They map phrase-structure parses onto SFL networks, but coverage is brittle outside the newswire sentences on which the rules were crafted. Recent graph-based pipelines transform dependency trees into reduced mood and transitivity networks \cite{costetchi2022graph}. Supervised approaches improve verb-class recognition provided that expert-annotated data exist \cite{yan2019process}. A recent survey concludes that SFL parsers remain “experimental, domain-sensitive, and labour-intensive to adapt” \cite{odonnell2016handbook}.  

Our framework departs from this rule-engineering paradigm. We prompt a large language model to jointly label processes, participants, and circumstances. This few-shot approach removes the need for handcrafted grammars or verb lexicons, and it scales to thousands of narratives without additional annotation effort.

\paragraph{Computational analysis of dream narratives.}
Quantitative dream research traditionally depends on manual coding systems, which is labour-intensive and difficult to scale. Rule‐ and lexicon‐based methods map dream words to hand-curated dictionaries or WordNet synsets to identify characters, interactions, or emotions \cite{Bulkeley2018,ourdreams,Mallett2021,Zheng2021}. Distributional approaches embed dreams in a semantic space and cluster them by themes such as flight or fight \cite{ALTSZYLER2017178,10.3389/fnins.2018.00007,GUTMANMUSIC2022103428}. Hybrid systems add classifiers on top of lexicon scores to predict overall sentiment or social themes \cite{McNamara_Duffy-Deno_Marsh_Marsh_2019,Yu2022}. Language-model approaches are only emerging: \citet{bertolini2023automatic} fine-tune BERT to detect the presence of emotions, and \citet{cortal-2024-sequence} fine-tune a sequence-to-sequence transformer to predict character and emotion. All of these approaches mainly focus on \textit{what} is said (\textit{e.g.}, entities, emotions, or thematic content). In contrast, we target \textit{how} subjective experience is constructed linguistically.

\section{Conclusion and Perspectives}

We present a novel framework that formalizes style in personal narratives as patterns in linguistic choices used to communicate subjective experience. By combining functional linguistics, sequence analysis methods, and psychological interpretations of sequential patterns, we analyze how authors encode their experiences through language. Our framework demonstrates the potential for uncovering psychologically meaningful patterns in dream narratives. The analysis of a war veteran reveals distinctive patterns in process types, particularly a prevalence of verbal processes over mental ones, suggesting links between linguistic choices and trauma processing. Our work opens directions for future research. 
\paragraph{Authorship profiling} Authorship profiling aims to infer an author's demographic and psychological characteristics based on their written texts. Applications range from forensic linguistics \cite{gibbonsForensicLinguisticsIntroduction2003} and literary attribution \cite{ledgerShakespeareFletcherTwo1994} to online security \cite{argamonAutomaticallyProfilingAuthor2009} and bot detection \cite{ferraraDetectionPromotedSocial2016}. Our framework could enhance existing approaches to authorship attribution, particularly for texts where patterns related to subjective experience diagnose authorship. We can identify signature patterns (\textit{e.g.}, distinctive substrings) that characterize an author's unique way of constructing narratives.

\paragraph{Style-conditioned narrative generation} Language models have demonstrated increasing utility in psychotherapeutic applications \cite{cortal-etal-2023-emotion,bonard-cortal-2024-improving}, particularly in text transformation tasks such as cognitive reframing of negative thoughts \cite{sharmaCognitiveReframingNegative2023b} and generating positive perspectives \cite{ziemsInducingPositivePerspectives2022a}. While most research in controllable text generation has focused on high-level attributes like formality and sentiment \cite{jinDeepLearningText2022,troianoTheoriesStylesTheir2023a}, recent work has shifted toward controlling fine-grained linguistic attributes \cite{alhafniPersonalizedTextGeneration2024b}. Our sequence-based framework offers a novel approach to controlling fine-grained attributes related to communicating subjective experience. While our current work focuses on structuring linguistic choices to characterize style from narratives, we could explore the inverse process: generating narratives from sequences of choices. This inverse mapping could enable style-conditioned narrative generation, offering new possibilities for therapeutic interventions or literary writing. 

\paragraph{Applying methods from complexity science and formal language theory} While our current analysis focuses on substrings, analyzing subsequences might offer more flexible pattern detection as it involves non-contiguous patterns. Moreover, we could draw from complexity measures to quantify redundancies in sequences. For example, the Lempel-Ziv measure evaluates how efficiently a sequence can be parsed into non-redundant substrings \cite{lempelComplexityFiniteSequences1976}, a low complexity indicating predictable patterns. These measures can reveal psychological insights: low complexity may indicate fixed mental representations or recurring cognitive schemas, particularly relevant in trauma cases, while higher complexity might suggest greater cognitive flexibility. This aligns with research applying complexity science to understand mental disorders \cite{carhart-harrisEntropicBrainRevisited2018,hipolitoPatternBreakingComplex2023a,hongLexicalUseEmotional2015}.

\section*{Limitations}

Our approach relies on language models to segment and annotate clauses. Although automation enables large-scale analysis, it can introduce inaccuracies when models misclassify processes or participants. Such inaccuracies can undermine the reliability of observed patterns. Improvements in model performance, as well as future efforts to validate or correct automated annotations using expert-annotated data, would help ensure more trustworthy results.

While we connect patterns in linguistic choices to psychological interpretations, these connections remain correlational and descriptive. Identifying a higher or lower prevalence of certain processes (\textit{e.g.}, mental or verbal) may hint at underlying psychological states, but these inferences are speculative. Validation against clinical assessments would be necessary to strengthen claims about the therapeutic utility of our stylistic analyses.


Despite these limitations, our approach lays the groundwork for a sequence-based, linguistically informed analysis of personal narratives. Future studies can address these limitations by expanding the range of linguistic features, refining extraction methods, and validating psychological interpretations with clinical data.

\section*{Ethics Statement}

Our study involves autobiographical dream narratives that may reveal mental‑health information about the dreamers. All texts are obtained from the publicly available DreamBank database and are anonymized. All dreamers are identified only by pseudonyms supplied by the original collectors. The dreamers have given their consent to appear in the database. No attempt was made to infer individual clinical diagnoses beyond the descriptive case study, and we emphasise that stylistic patterns should never be treated as diagnostic in isolation.

The automatic annotations are produced with an open‑weights language model (Llama 3.1 8B). Because such models can encode socially biased associations, we release the prompts, hyperparameters and extracted sequences to enable third‑party auditing\footnote{\url{https://github.com/gustavecortal/formalizing-style-in-personal-narratives}}. Potential downstream uses (\textit{e.g.}, automated triage) might amplify misclassification harms. We stress that our framework is intended for exploratory research or therapist‑in‑the‑loop settings, not fully autonomous clinical deployment.

\section*{Acknowledgments}

Partially supported by the SAIF project, funded by the “France 2030” government investment plan and managed by the ANR under the reference ANR-23-PEIA-0006. This work was performed using HPC resources from GENCI-IDRIS (Grant 20XX-AD011014205).

\bibliography{custom,lrec-coling2024-example}


\appendix

\section{Appendix}
\label{sec:appendix}

In the next page, we describe processes and participants. We also provide more statistical analysis of sequences such as the number of distinct substrings in the series, top substrings for each series, and the distribution of sequence lengths.

\begin{table*}
    \centering
    \begin{tabular}{p{7cm}|p{8cm}}
        \hline
        \textbf{Processes} & \textbf{Participants} \\ \hline
        \textit{Action}: actions and events in the physical world. &
        \textit{Actor}: performs the action \newline
        \textit{Affected}: receives the impact of the action \newline
        \textit{Result}: outcome or creation from the process \newline
        \textit{Recipient}: entity to whom the action is directed \\ \hline
        
        \textit{Mental}: internal experiences such as thoughts, perceptions, and feelings. &
        \textit{Senser}: experiencer of the process \newline
        \textit{Phenomenon}: what is being experienced \\ \hline
        
        \textit{Verbal}: acts of communication. &
        \textit{Sayer}: communicator of the message \newline
        \textit{Verbiage}: content of the message \newline
        \textit{Recipient}: receiver of the message \\ \hline
        

        \textit{States}: states of existence, being or having. &
        \textit{Existent}: entity that exists \newline
        \textit{Carrier}: entity linked to an \textit{Attribute} \newline
        \textit{Attribute}: quality or descriptor of the \textit{Carrier} \newline
        \textit{Possessor}: owner of something \newline
        \textit{Possessed}: item owned by the \textit{Possessor} \\ 
    \end{tabular}
    \caption{Processes and their participants.}
    \label{tab:process_types_no_examples}
\end{table*}

\begin{table*}
    \centering
    \begin{tabular}{p{4cm}|p{11cm}}
        \hline
        \textbf{Circumstances} & \textbf{Examples} \\ \hline
        
        \textit{Time}: When? & 
        She arrived \textbf{yesterday}. \newline 
        He visits his grandmother \textbf{every week}. \newline 
        They worked \textbf{for three hours}. \\ \hline

        \textit{Place}: Where? & 
        The meeting will be held \textbf{in the conference room}. \newline 
        They walked \textbf{to the park}. \newline 
        He came \textbf{from New York}. \\ \hline

        \textit{Manner}: How? & 
        She spoke \textbf{softly}. \newline 
        He fixed the car \textbf{with a wrench}. \newline 
        He loves her \textbf{very much}. \\ \hline

        \textit{Cause}: Why? & 
        She stayed home \textbf{because of the rai}n. \newline 
        He went to the store \textbf{to buy milk}. \\ \hline

        \textit{Accompaniment}: With whom? & 
        She went to the movie \textbf{with her friends}. \newline 
        He completed the project \textbf{without any help}. \\ \hline

        \textit{Matter}: About what? & 
        They had a discussion \textbf{about the new policy}. \newline 
        She wrote a report \textbf{on climate change}. \\ \hline

        \textit{Role}: As what? & 
        She attended the meeting \textbf{as a representative}. \newline 
        He acted \textbf{as the spokesperson}. \\ 
    \end{tabular}
    \caption{Types of circumstances with examples.}
    \label{tab:circumstances}
\end{table*}

\begin{table}
    \centering
    \label{tab:my_label}
    \begin{tabular}{llll}
        \hline
        \textbf{Series} & \textbf{Size} & {\textbf{Count}} & {\textbf{Sig. Diff.}} \\
        \hline
    \multirow{9}{*}{\textit{viet}} 
          & 1 & 4    & 3  \\
          & 2 & 16   & 13 \\
          & 3 & 64   & 35 \\
          & 4 & 253  & 59 \\
          & 5 & 864  & 73 \\
          & 6 & 2017 & 68 \\
          & 7 & 3337 & 42 \\
          & 8 & 4283 & 23 \\
          & 9 & 4773 & 13 \\
    \hline
    \multirow{9}{*}{\textit{blind}} 
          & 1 & 4    & 3  \\
          & 2 & 16   & 5  \\
          & 3 & 64   & 12 \\
          & 4 & 253  & 27 \\
          & 5 & 786  & 55 \\
          & 6 & 1570 & 47 \\
          & 7 & 2137 & 33 \\
          & 8 & 2285 & 8  \\
          & 9 & 2202 & 6  \\
    \hline
    \multirow{9}{*}{\textit{izzy}} 
          & 1 & 4    & 3  \\
          & 2 & 16   & 12 \\
          & 3 & 64   & 39 \\
          & 4 & 255  & 79 \\
          & 5 & 930  & 101\\
          & 6 & 2520 & 98 \\
          & 7 & 4828 & 92 \\
          & 8 & 7007 & 58 \\
          & 9 & 8568 & 26 \\
    \hline
    \multirow{9}{*}{\textit{ed}} 
          & 1 & 4    & 4  \\
          & 2 & 16   & 8  \\
          & 3 & 64   & 26 \\
          & 4 & 254  & 47 \\
          & 5 & 783  & 73 \\
          & 6 & 1406 & 93 \\
          & 7 & 1722 & 67 \\
          & 8 & 1790 & 39 \\
          & 9 & 1727 & 21 \\
    \hline
    \multirow{9}{*}{\textit{merri}} 
          & 1 & 4    & 2  \\
          & 2 & 16   & 9  \\
          & 3 & 64   & 21 \\
          & 4 & 246  & 23 \\
          & 5 & 740  & 22 \\
          & 6 & 1564 & 40 \\
          & 7 & 2447 & 41 \\
          & 8 & 3157 & 23 \\
          & 9 & 3610 & 17 \\
    \end{tabular}
        \caption{\textbf{Series} is the series name, \textbf{Size} is the substring size, \textbf{Count} is the number of distinct substrings in the series, \textbf{Sig. Diff.} is the count of substrings with significant differences relative to the norm.}
        \label{fig:distinct_substrings}
\end{table}

\begin{figure}
     \centering
     \begin{subfigure}[t]{\linewidth}
         \centering
         \includegraphics[scale=0.28]{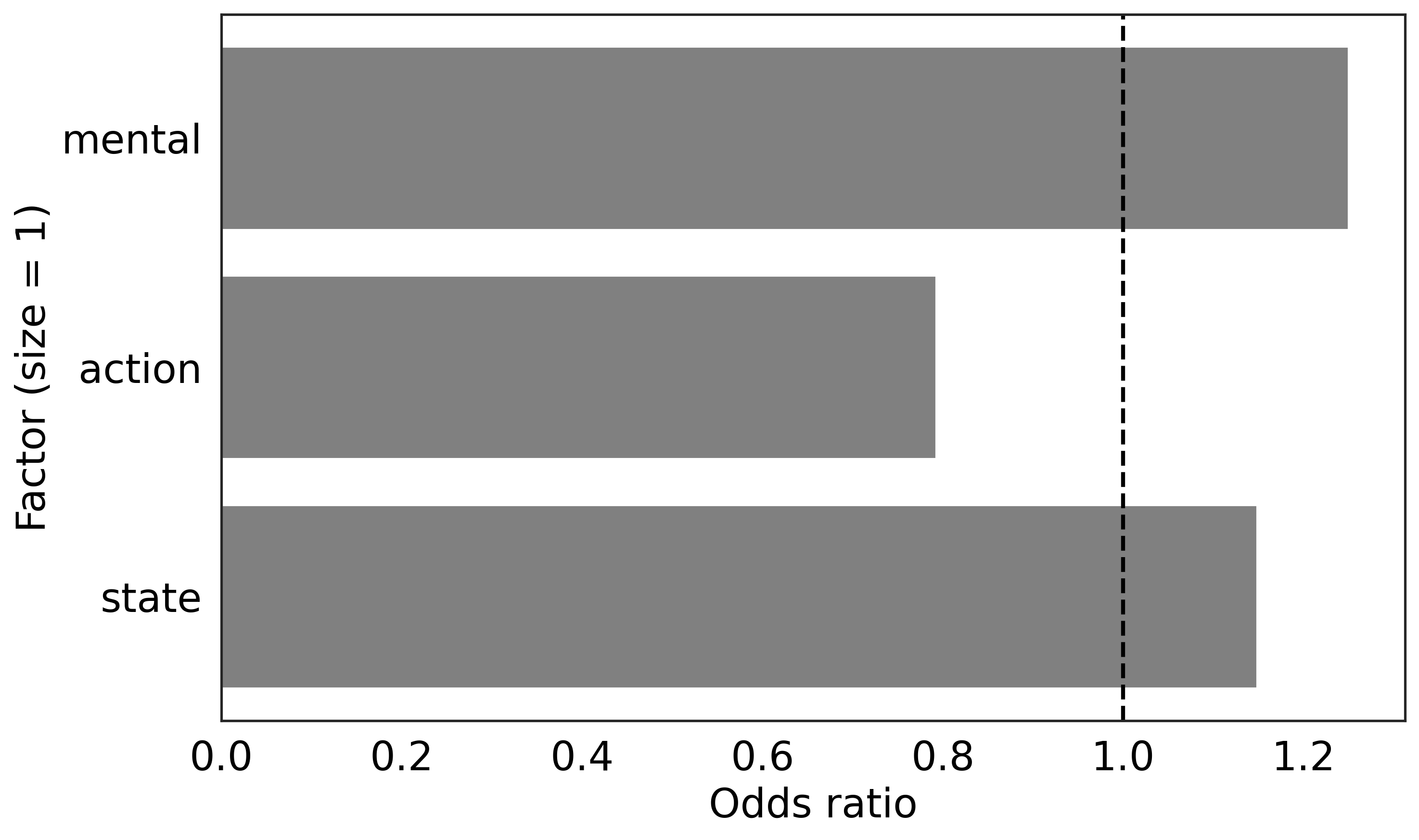}
         \caption{Size 1.}
     \end{subfigure}
     \medskip
     \begin{subfigure}[t]{\linewidth}
         \centering
         \includegraphics[scale=0.28]{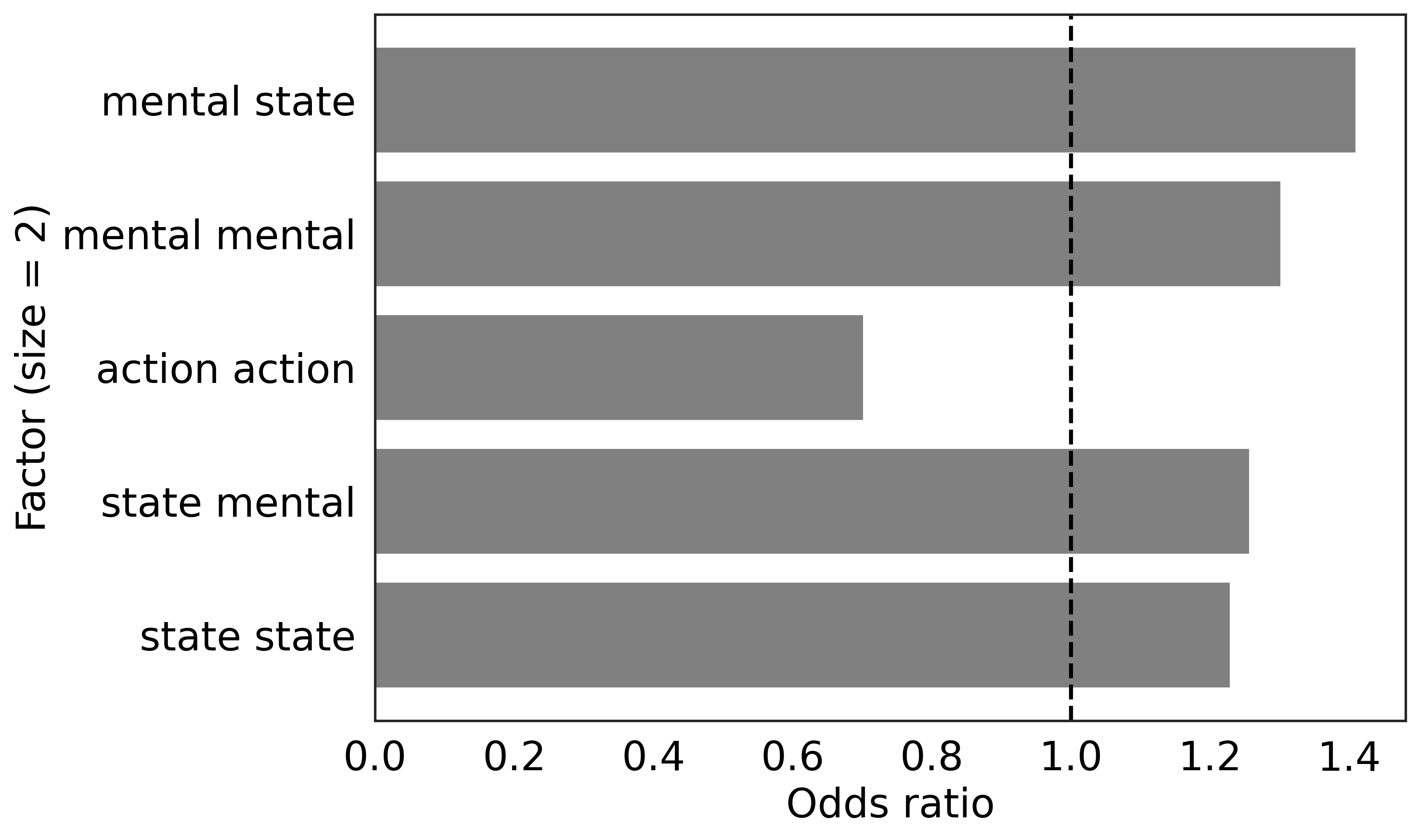}
         \caption{Size 2.}
     \end{subfigure}
         \begin{subfigure}[t]{\linewidth}
         \centering
         \includegraphics[scale=0.28]{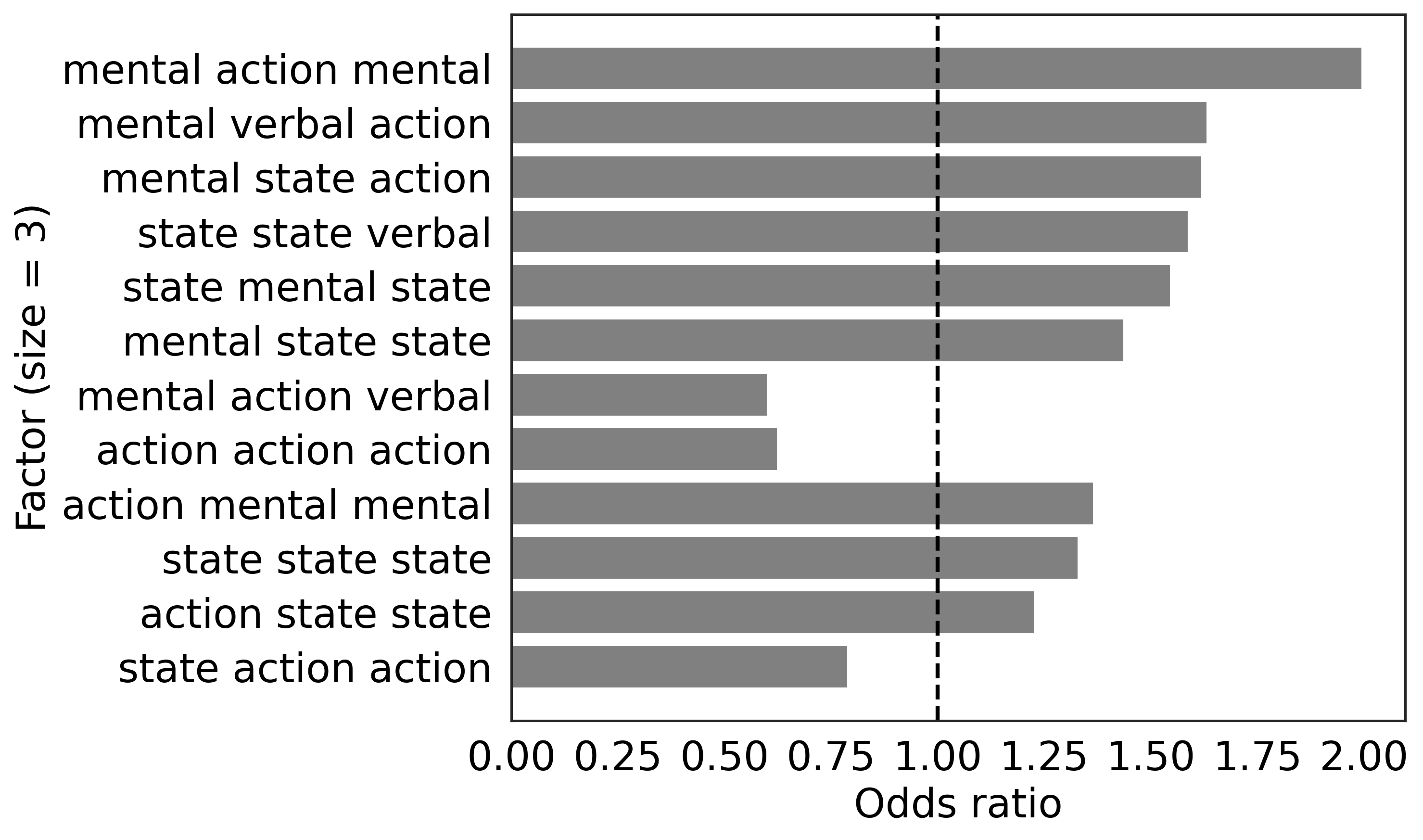}
         \caption{Size 3.}
     \end{subfigure}
     \medskip
     \begin{subfigure}[t]{\linewidth}
         \centering
         \includegraphics[scale=0.28]{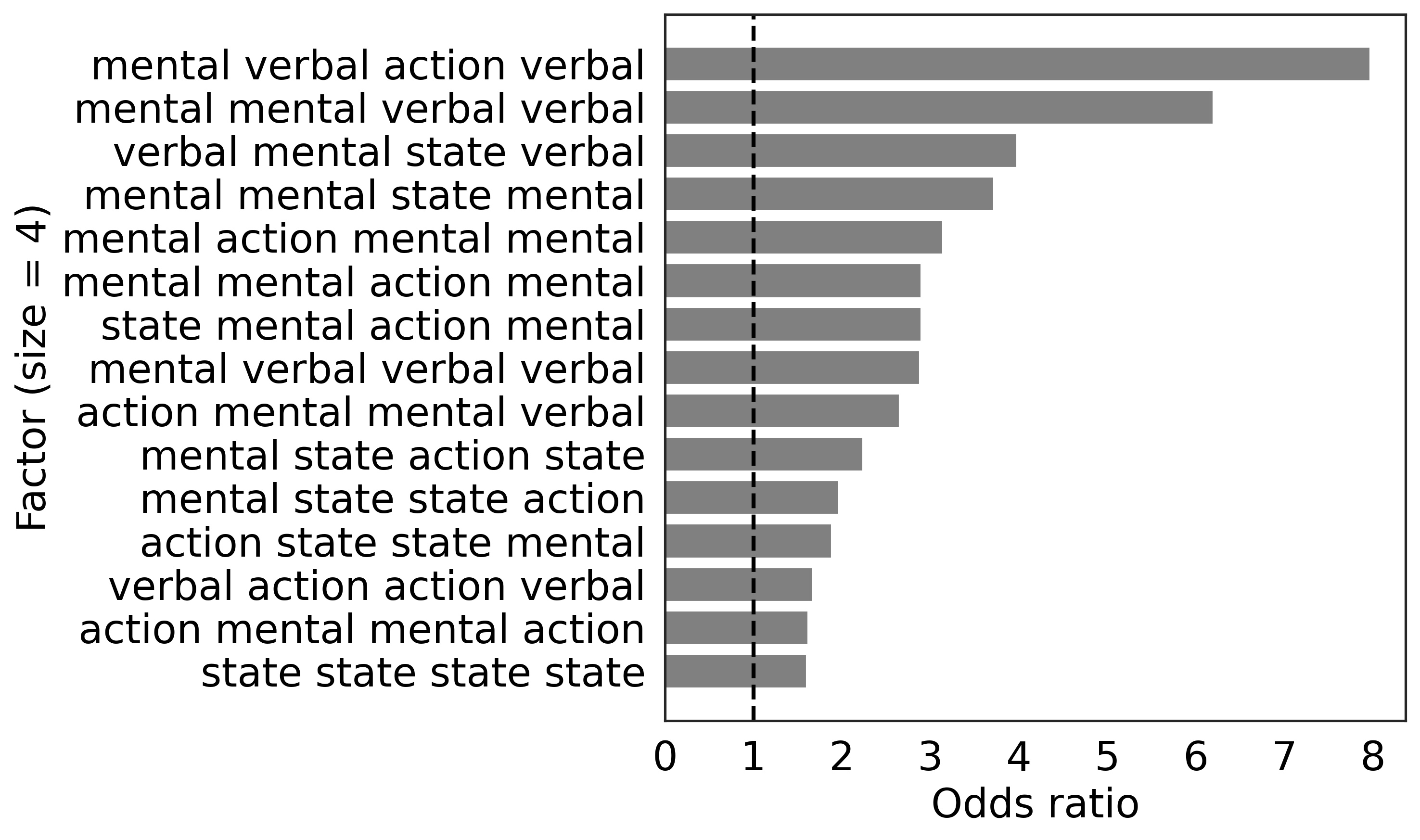}
         \caption{Size 4.}
     \end{subfigure}
        \caption{Top substring between \textit{blind} and \textit{norm}.}
\end{figure}

\begin{figure}
     \centering
     \begin{subfigure}[t]{\linewidth}
         \centering
         \includegraphics[scale=0.28]{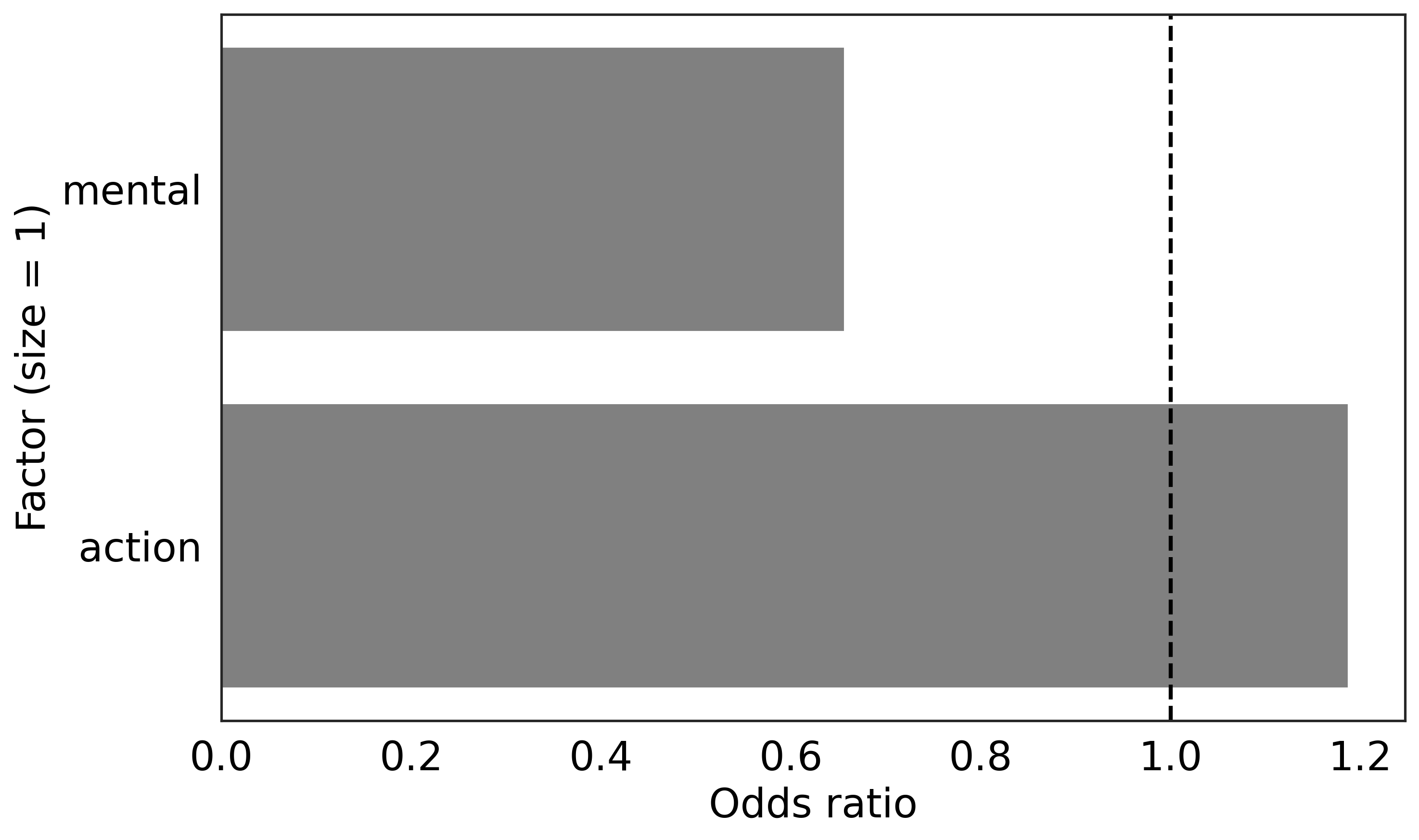}
         \caption{Size 1.}
     \end{subfigure}
     \medskip
     \begin{subfigure}[t]{\linewidth}
         \centering
         \includegraphics[scale=0.28]{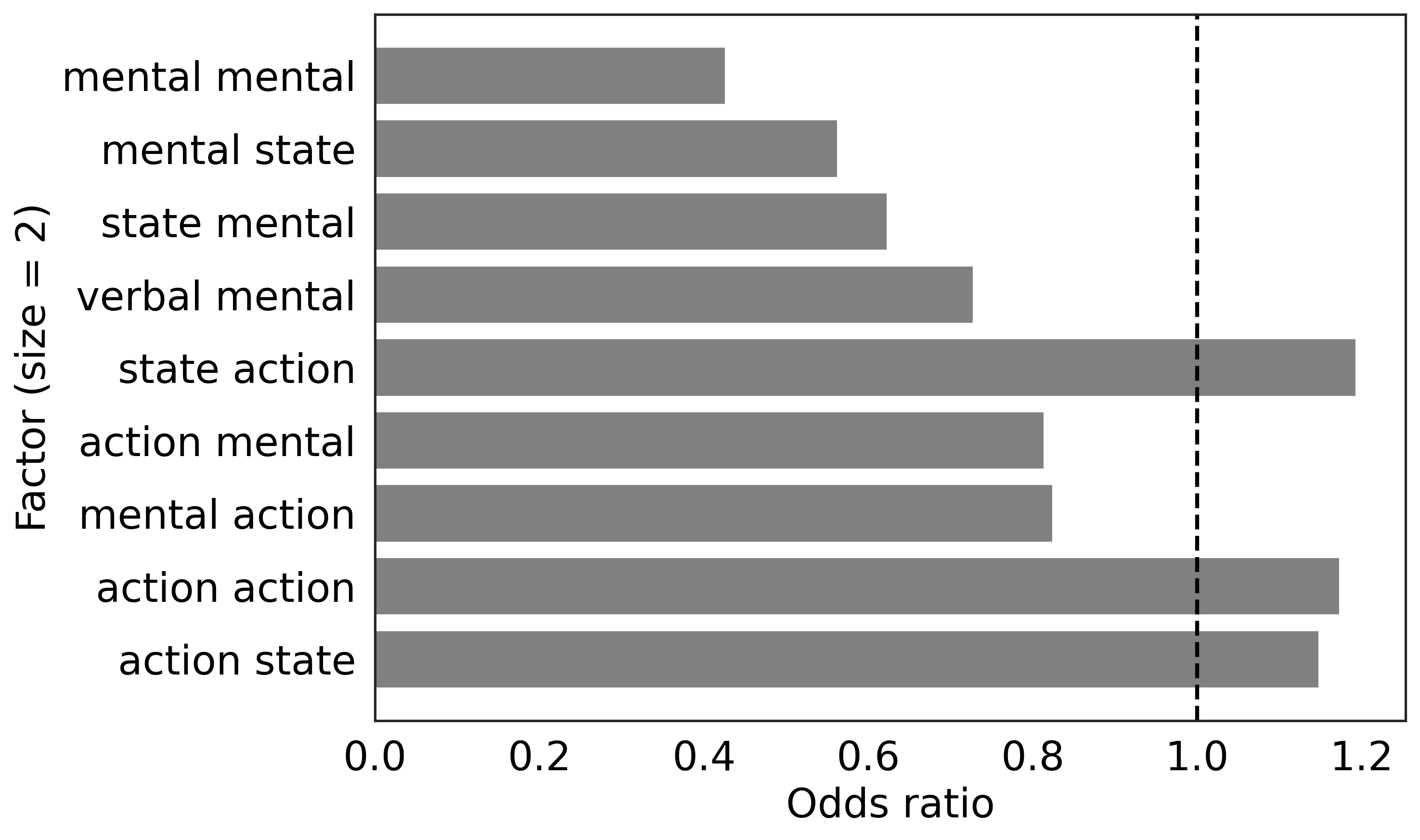}
         \caption{Size 2.}
     \end{subfigure}
         \begin{subfigure}[t]{\linewidth}
         \centering
         \includegraphics[scale=0.28]{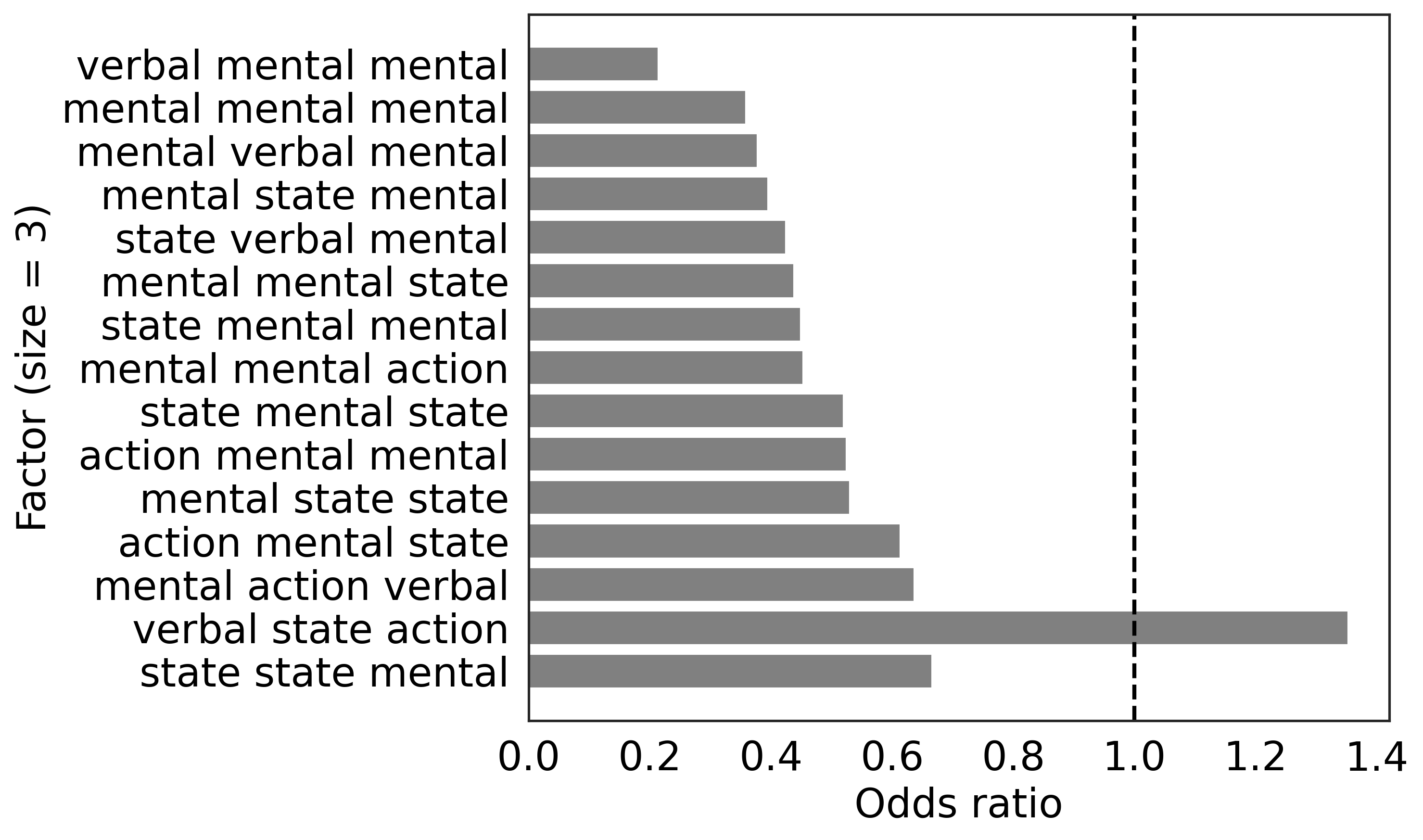}
         \caption{Size 3.}
     \end{subfigure}
     \medskip
     \begin{subfigure}[t]{\linewidth}
         \centering
         \includegraphics[scale=0.28]{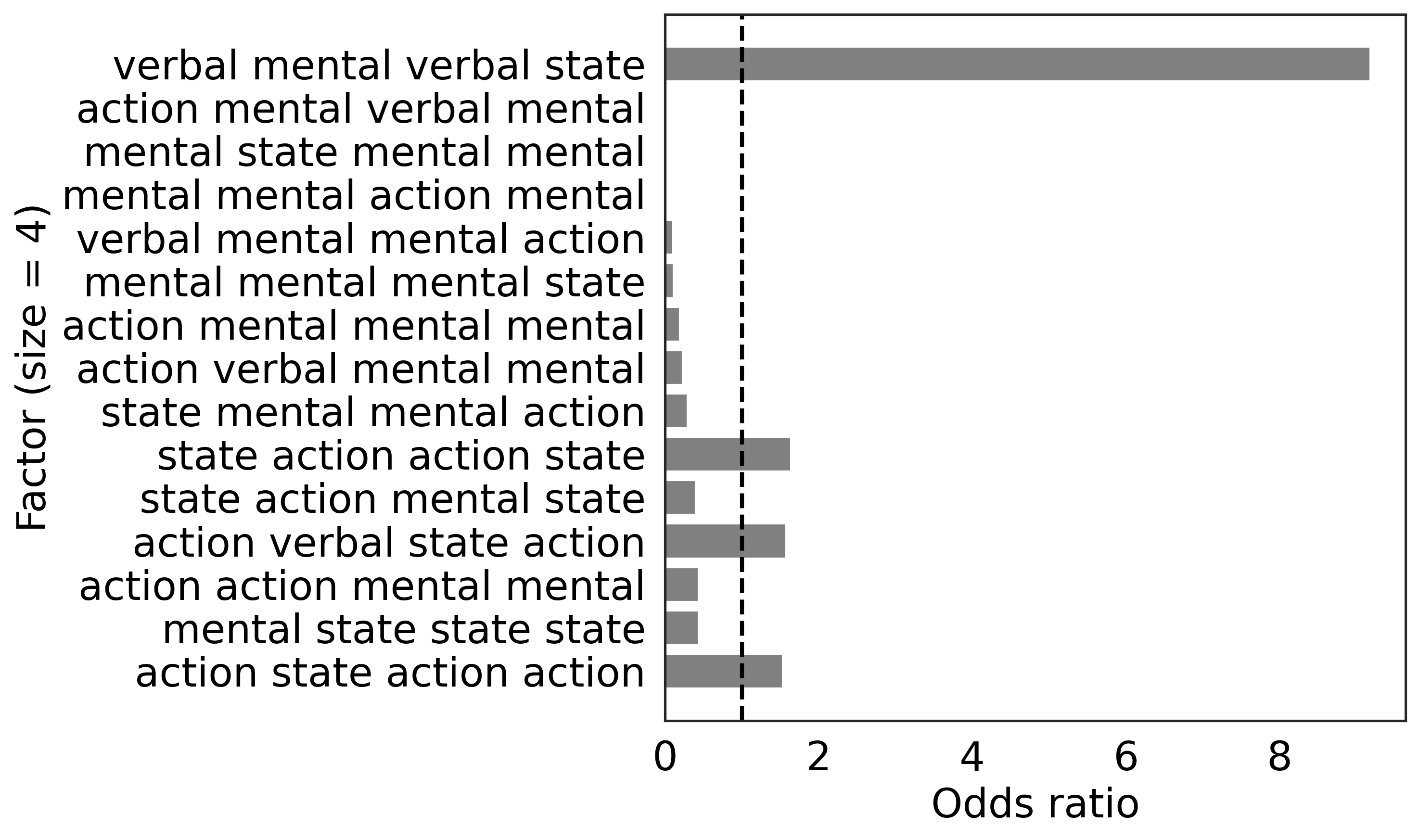}
         \caption{Size 4.}
     \end{subfigure}
        \caption{Top substring between \textit{merri} and \textit{norm}.}
\end{figure}

\begin{figure}
     \centering
     \begin{subfigure}[t]{\linewidth}
         \centering
         \includegraphics[scale=0.28]{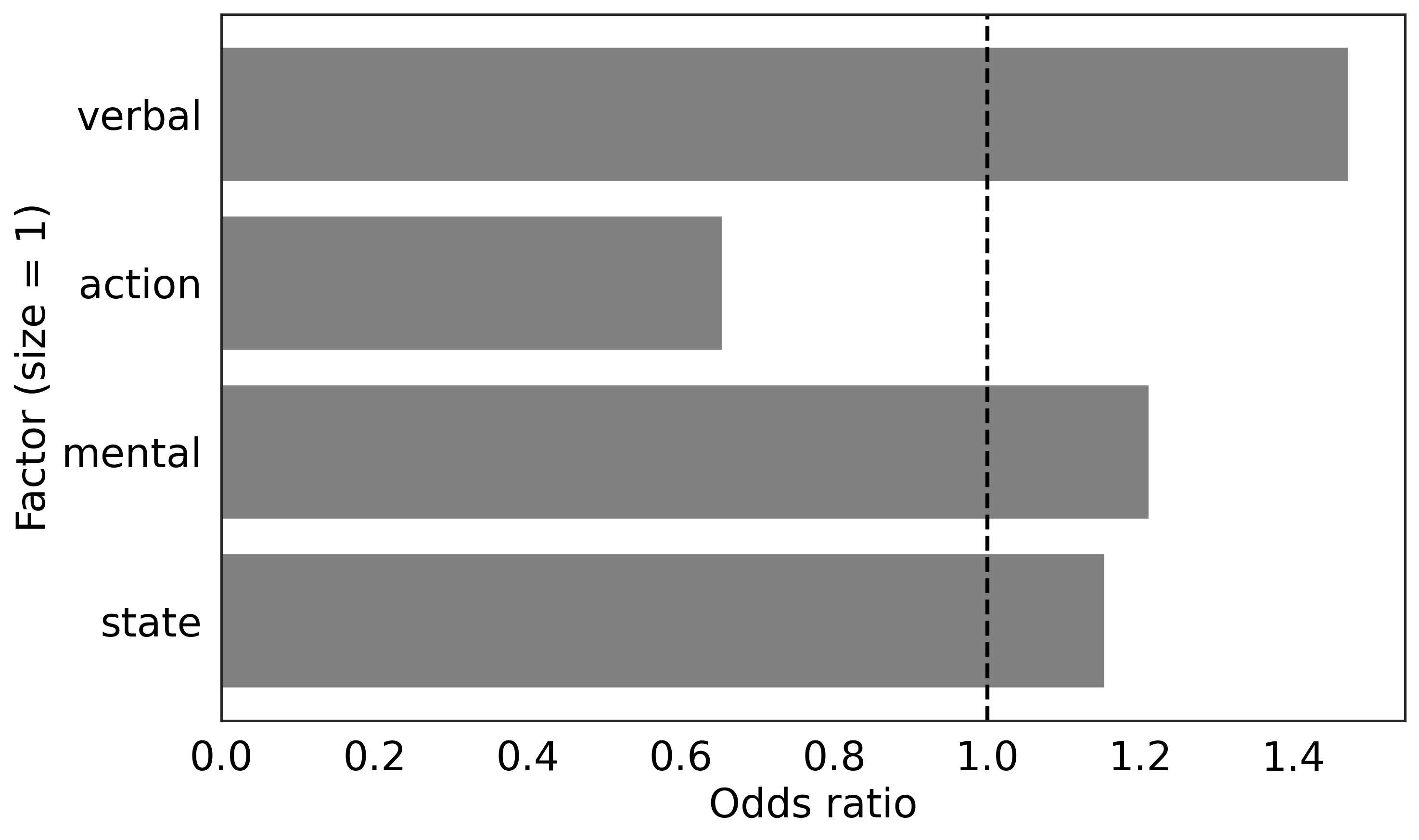}
         \caption{Size 1.}
     \end{subfigure}
     \medskip
     \begin{subfigure}[t]{\linewidth}
         \centering
         \includegraphics[scale=0.28]{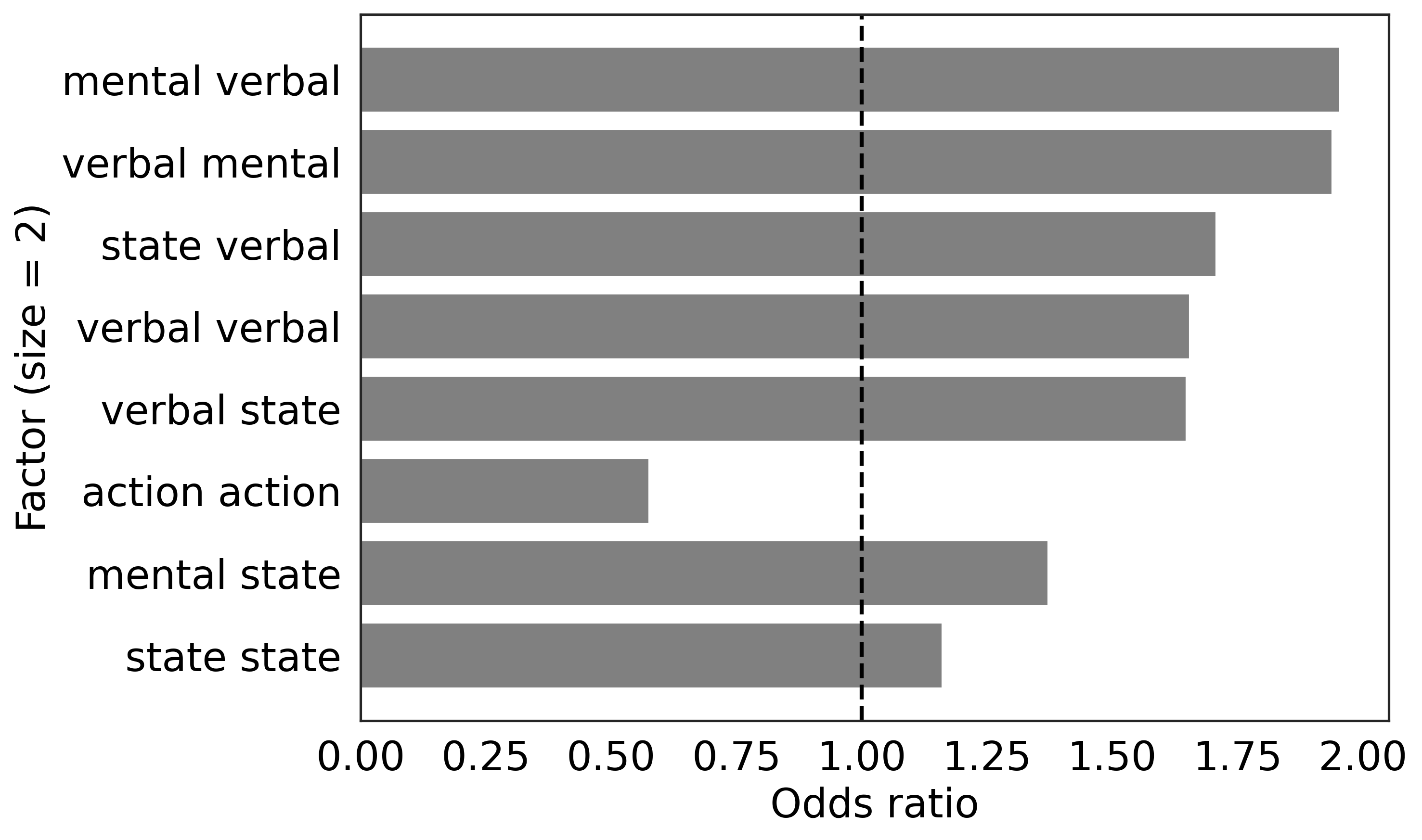}
         \caption{Size 2.}
     \end{subfigure}
         \begin{subfigure}[t]{\linewidth}
         \centering
         \includegraphics[scale=0.28]{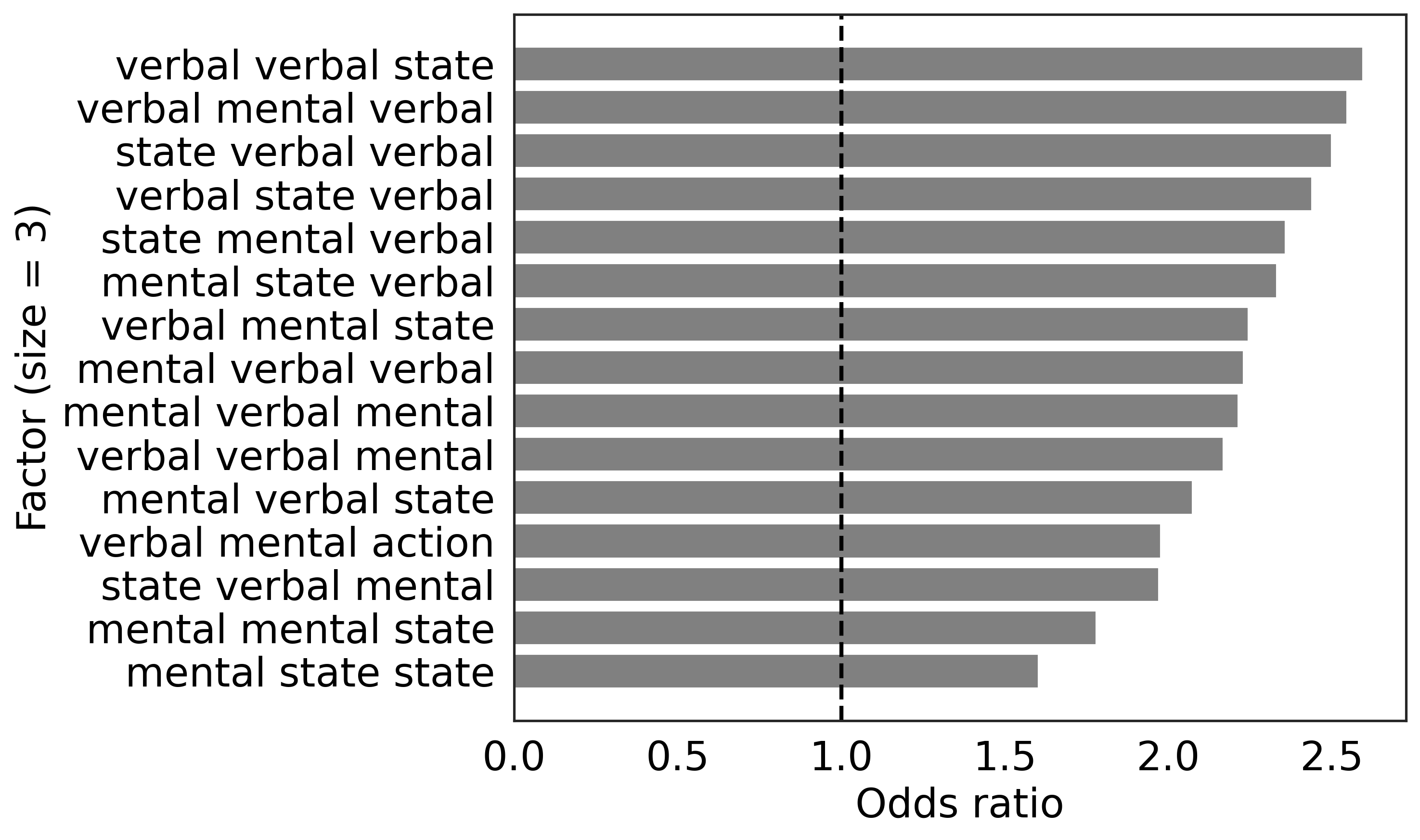}
         \caption{Size 3.}
     \end{subfigure}
     \medskip
     \begin{subfigure}[t]{\linewidth}
         \centering
         \includegraphics[scale=0.28]{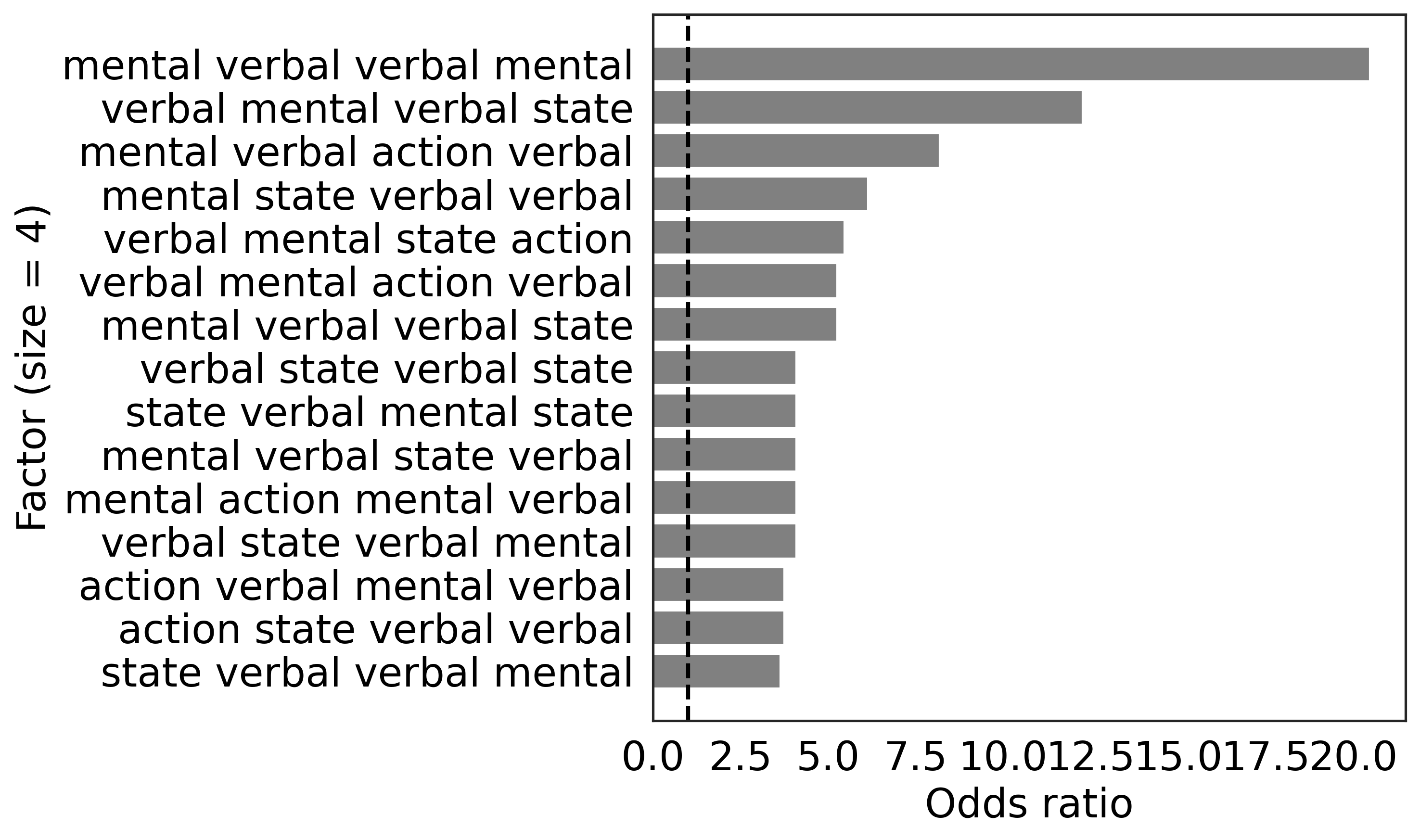}
         \caption{Size 4.}
     \end{subfigure}
        \caption{Top substring odds ratio between \textit{ed} and \textit{norm}.}
\end{figure}

\begin{figure}
     \centering
     \begin{subfigure}[t]{\linewidth}
         \centering
         \includegraphics[scale=0.28]{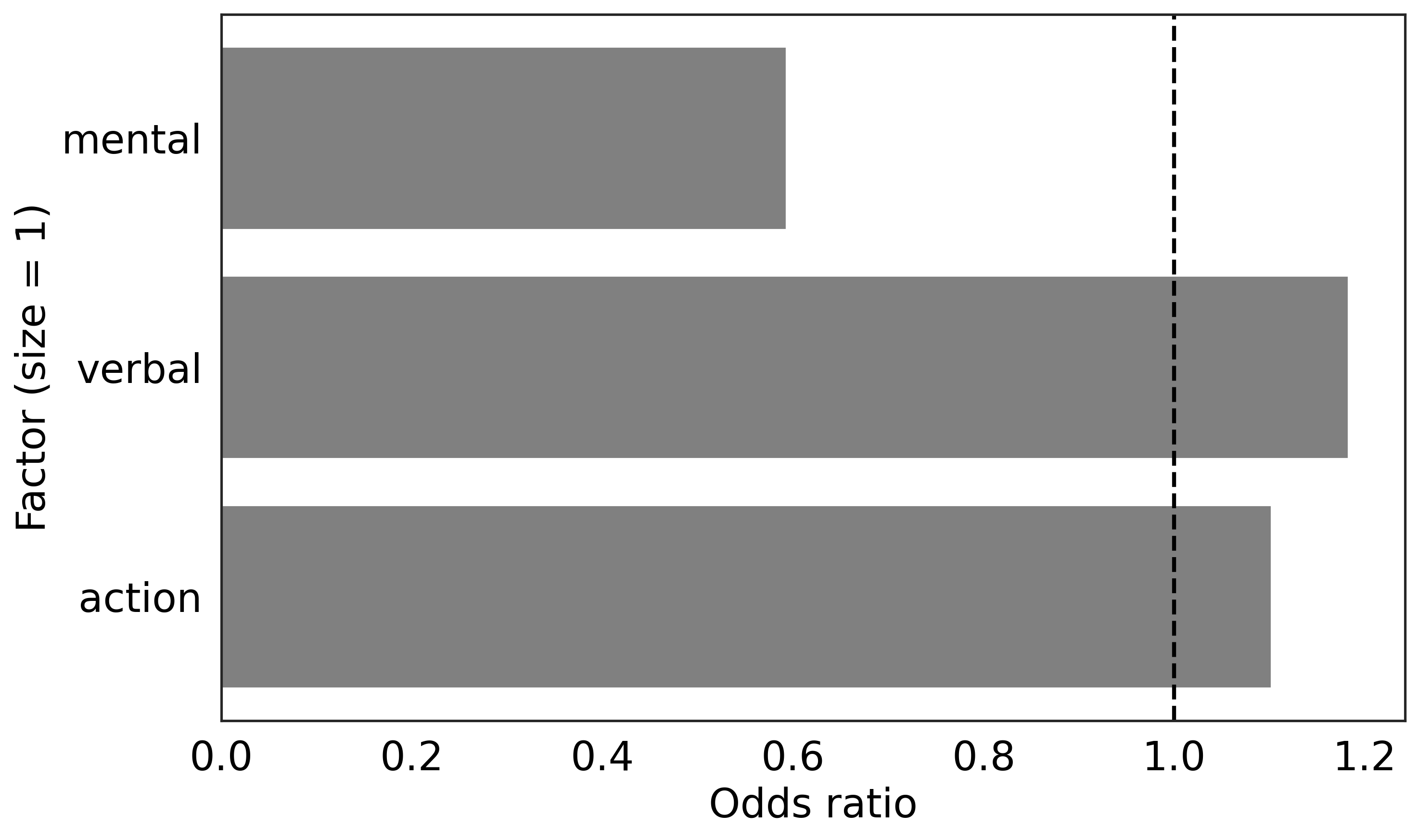}
         \caption{Size 1.}
     \end{subfigure}
     \medskip
     \begin{subfigure}[t]{\linewidth}
         \centering
         \includegraphics[scale=0.28]{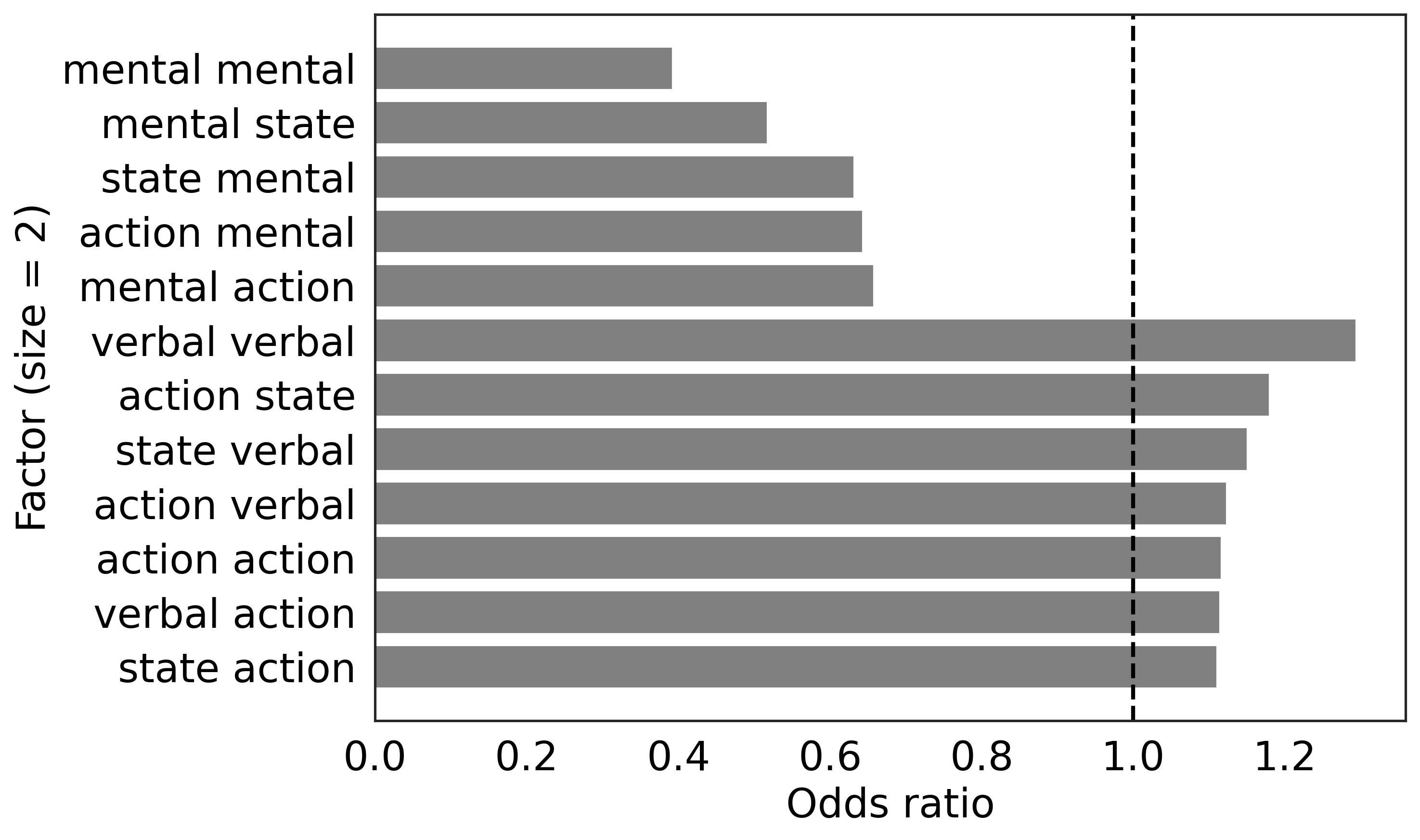}
         \caption{Size 2.}
     \end{subfigure}
         \begin{subfigure}[t]{\linewidth}
         \centering
         \includegraphics[scale=0.28]{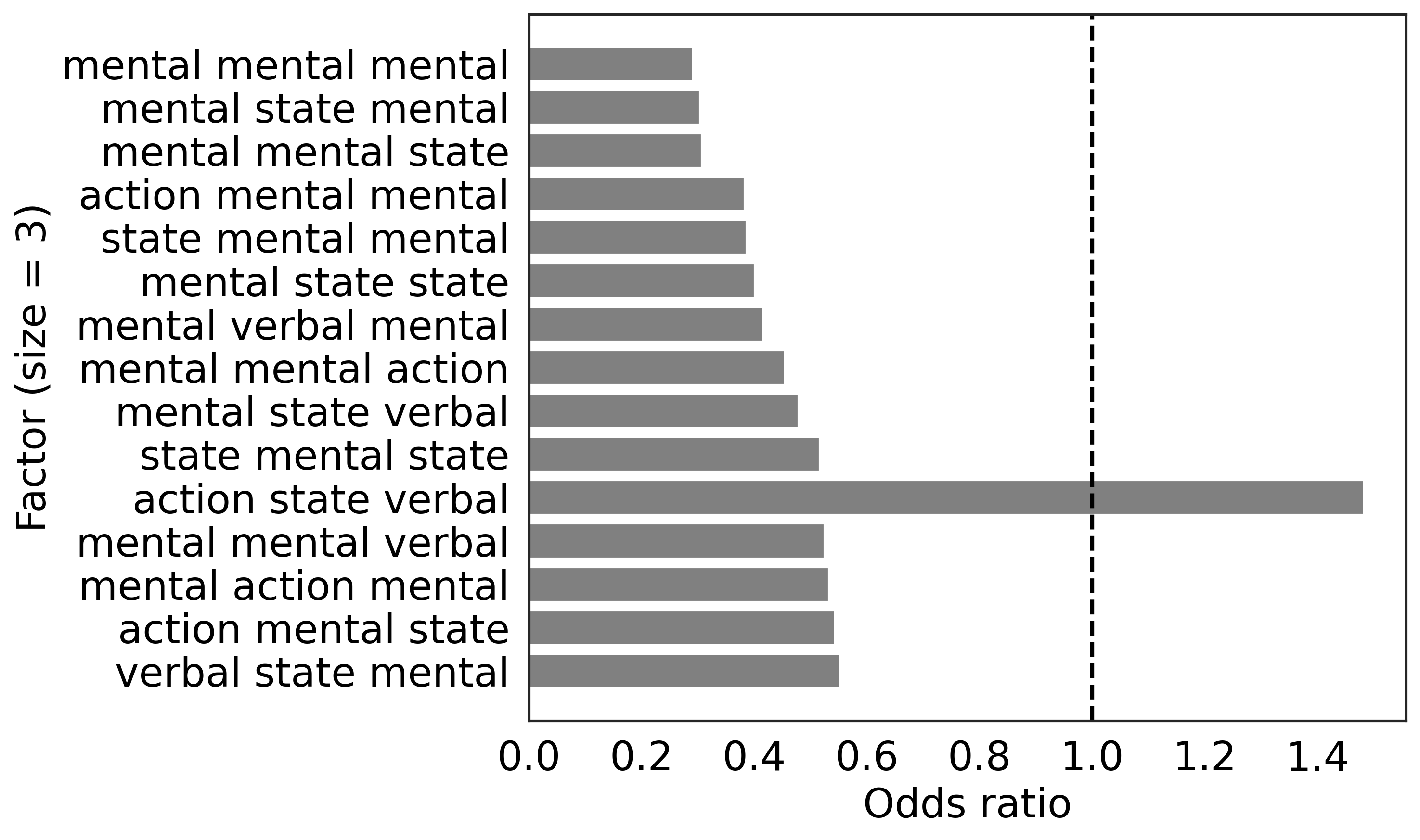}
         \caption{Size 3.}
     \end{subfigure}
     \medskip
     \begin{subfigure}[t]{\linewidth}
         \centering
         \includegraphics[scale=0.28]{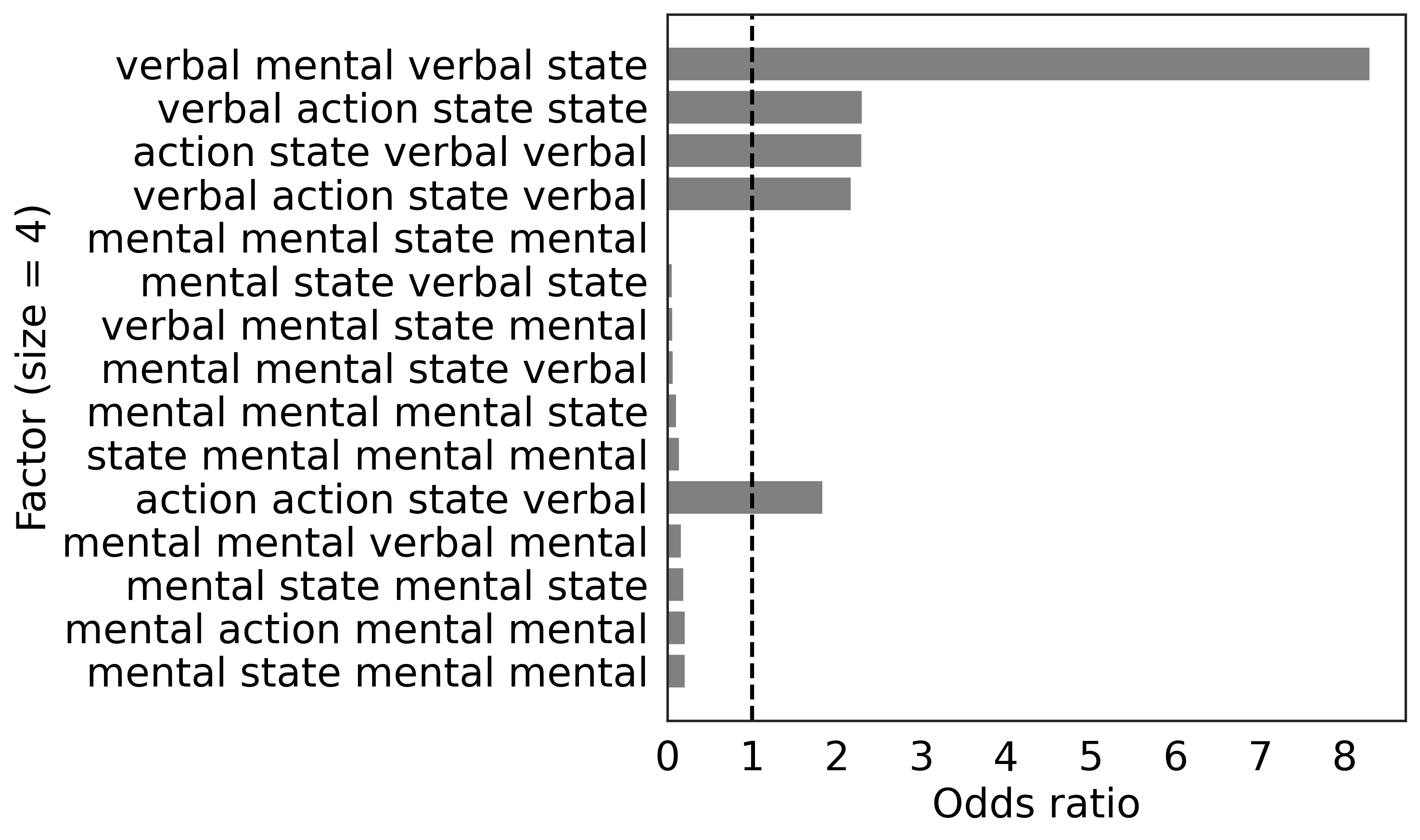}
         \caption{Size 4.}
     \end{subfigure}
        \caption{Top substring between \textit{izzy} and \textit{norm}.}
\end{figure}

\begin{figure}[ht]
     \centering
     \begin{subfigure}[t]{\linewidth}
         \centering
         \includegraphics[scale=0.28]{latex/images_subj_exp/viet_odds_1.png}
         \caption{Size 1.}
     \end{subfigure}
     \medskip
     \begin{subfigure}[t]{\linewidth}
         \centering
         \includegraphics[scale=0.28]{latex/images_subj_exp/viet_odds_2.png}
         \caption{Size 2.}
     \end{subfigure}
         \begin{subfigure}[t]{\linewidth}
         \centering
         \includegraphics[scale=0.28]{latex/images_subj_exp/viet_odds_3.png}
         \caption{Size 3.}
     \end{subfigure}
     \medskip
     \begin{subfigure}[t]{\linewidth}
         \centering
         \includegraphics[scale=0.28]{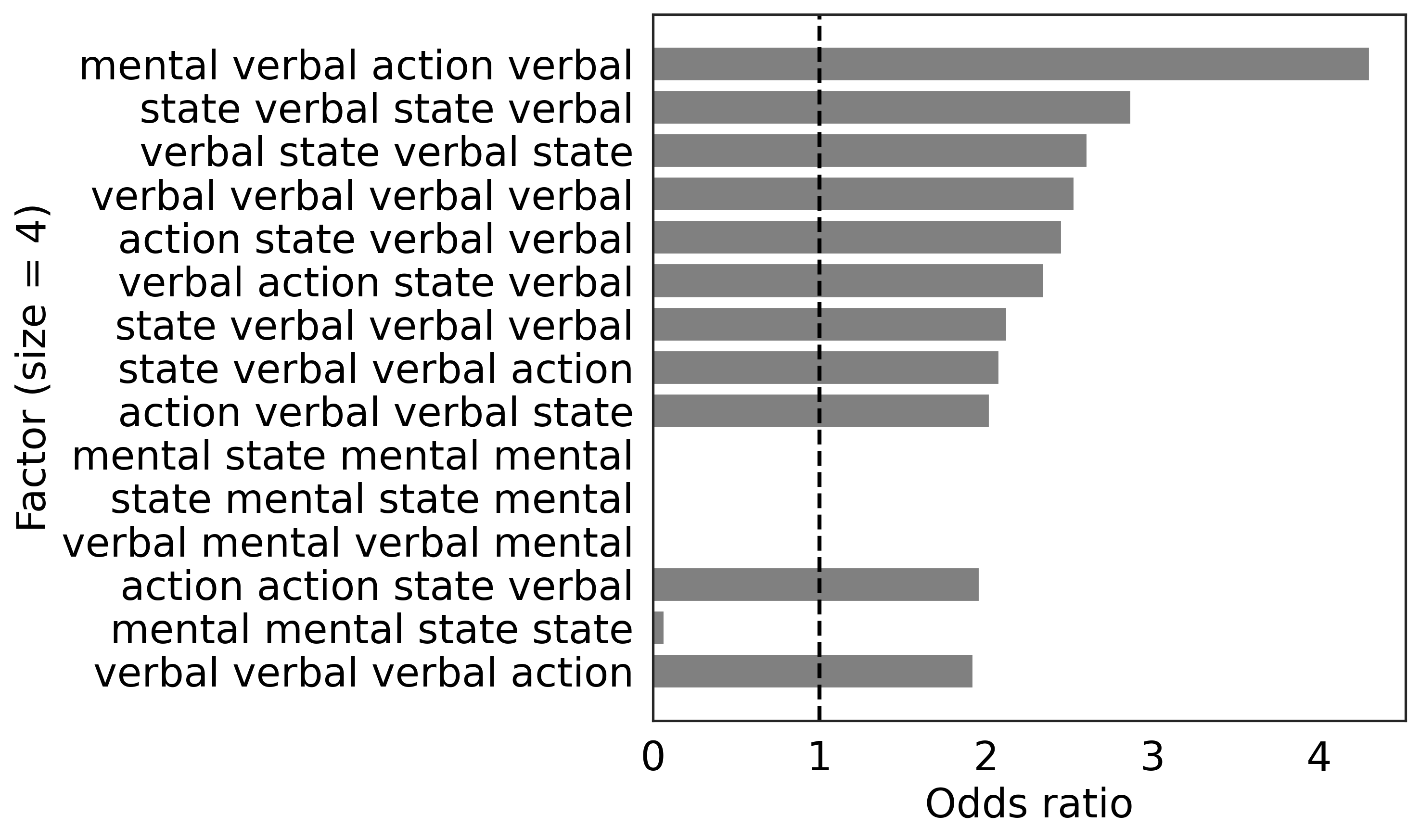}
         \caption{Size 4.}
     \end{subfigure}
        \caption{Top substring between \textit{viet} and \textit{norm}.}
\end{figure}

\begin{figure*}
    \centering
    \includegraphics[scale=0.4]{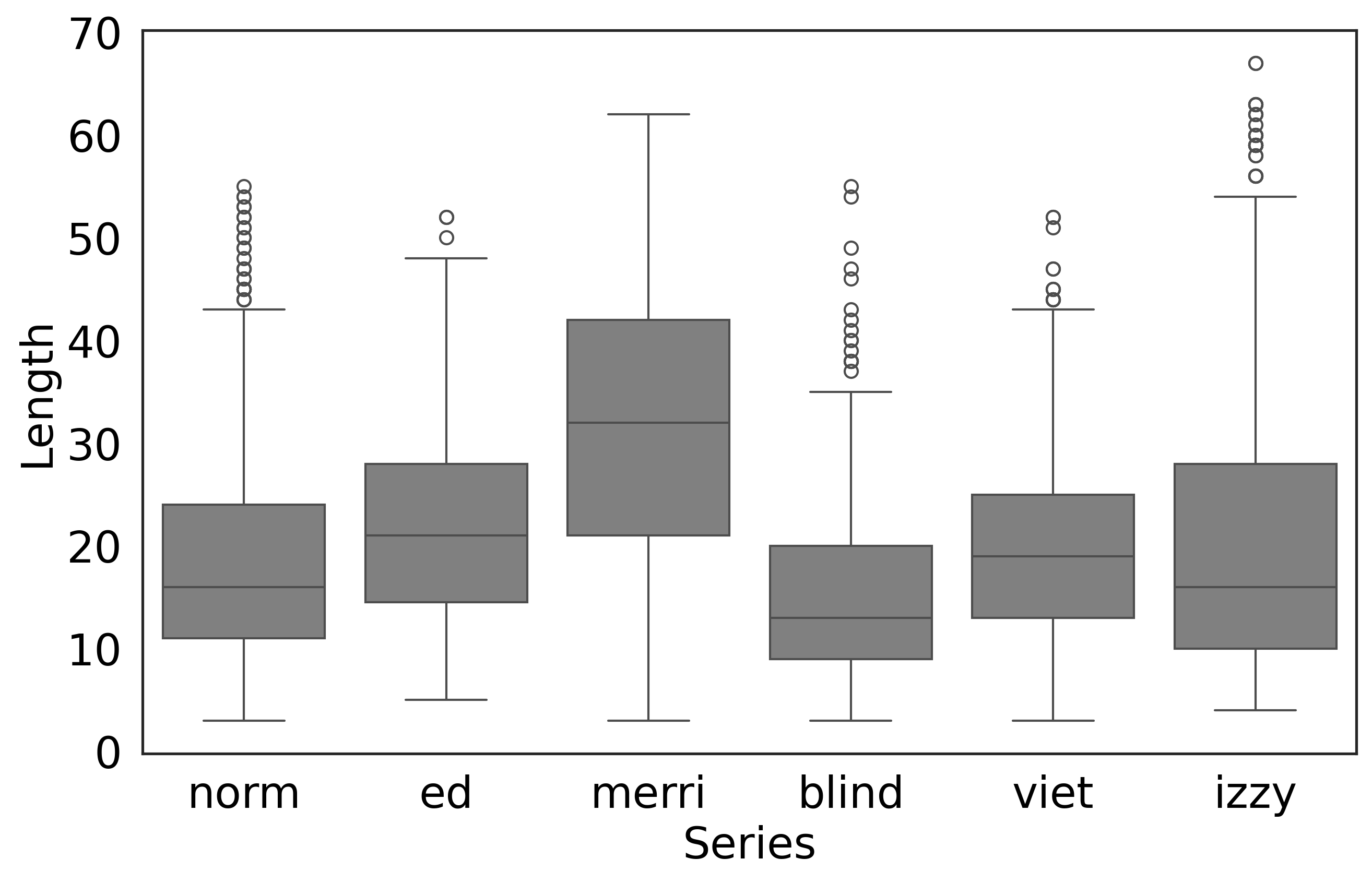}
    \caption{Per-series distribution of sequence lengths.}
    \label{fig:sequence_length}
\end{figure*}


\end{document}